\documentclass{article}

\usepackage{microtype}
\usepackage{graphicx}
\usepackage{booktabs} 
\usepackage{caption}
\usepackage{subcaption}

\usepackage{amsmath}
\usepackage{amsfonts}

\usepackage{hyperref}

\usepackage[accepted]{icml2020}

\icmltitlerunning{Combining Differentiable PDE Solvers and Graph Neural Networks for Fluid Flow Prediction}

\begin{document}

\twocolumn[
\icmltitle{Combining Differentiable PDE Solvers and \\ Graph Neural Networks for Fluid Flow Prediction}

\icmlsetsymbol{equal}{\dag}

\begin{icmlauthorlist}
\icmlauthor{Filipe de Avila Belbute-Peres}{cmu,equal}
\icmlauthor{Thomas D. Economon}{su2,equal}
\icmlauthor{J. Zico Kolter}{cmu,bosch}
\end{icmlauthorlist}

\icmlaffiliation{cmu}{School of Computer Science, Carnegie Mellon University, Pittsburgh, PA}
\icmlaffiliation{bosch}{Bosch Center for AI, Pittsburgh, PA}
\icmlaffiliation{su2}{SU2 Foundation}

\icmlcorrespondingauthor{Filipe de Avila Belbute-Peres}{\mbox{filiped@cs.cmu.edu}}

\icmlkeywords{Machine Learning, ICML, graph, mesh, convolution, CFD, fluid, PDE, hybrid, deep learning}

\vskip 0.3in
]

\printAffiliationsAndNotice{\textsuperscript{\dag}Work partly done while at Bosch LLC.\\}

\begin{abstract}
Solving large complex partial differential equations (PDEs), such as those that arise in computational fluid dynamics (CFD), is a computationally expensive process.  This has motivated the use of deep learning approaches to approximate the PDE solutions, yet the simulation results predicted from these approaches typically do not generalize well to truly novel scenarios.  In this work, we develop a hybrid (graph) neural network that combines a traditional graph convolutional network with an embedded \emph{differentiable} fluid dynamics simulator inside the network itself.  By combining an actual CFD simulator (run on a much coarser resolution representation of the problem) with the graph network, we show that we can both generalize well to new situations and benefit from the substantial speedup of neural network CFD predictions, while also substantially outperforming the coarse CFD simulation alone.

\end{abstract}

\section{Introduction}
\label{introduction}

In recent years, the many empirical successes of deep learning have motivated scientists to explore its application in other areas, such as predicting the evolution of physical systems.  In this context, several recent papers have explored the application of deep models to approximate the solutions to partial differential equations (PDEs), particularly in the context of simulating fluid dynamics~\citep{afshar_prediction_2019, guo2016convolutional, wiewel_latent-space_2018, um_liquid_2017}. The behavior of fluids is a well-studied problem in the physical sciences, and predicting their dynamics involves solving the nonlinear Navier-Stokes PDEs. 
In order to perform computational fluid dynamics (CFD) simulations, these equations must be solved numerically.
One of the primary bottlenecks in more accurate and advanced CFD simulation is specifically the time it takes to run these models. 
It is not uncommon for a single simulation to take many days to weeks on massive supercomputing infrastructure.  Especially for the cases where fast iteration time is desired, for example when iterating over different aerodynamic design prototypes for a structure, then faster, learning-based surrogate models have spawned a great deal of interest.  However, despite the recent enthusiasm for this area, most deep learning models cannot capture the full complexity of the underlying equations, and, as we demonstrate in this work, can quickly start to produce poor results when testing on settings well outside the domain of their training data.

In this paper, we explore a hybrid approach that combines the benefits of (graph) neural networks for fast predictions, with the physical realism of an industry-grade CFD simulator.  Our system has two main components.  First, we construct a graph convolution network~\cite{kipf2016semi} (GCN), which operates directly upon the non-uniform mesh used in typical CFD simulation tasks. This use of GCNs is crucial because all realistic CFD solvers operate on these unstructured meshes rather than directly on the regular grid used by most prior work, which has typically used convolutional networks to approximate CFD simulations.  Second, and more fundamentally, we embed a (differentiable) CFD solver, operating on a much coarser resolution, \emph{directly} into the GCN itself.  Although typically treated as black-boxes, modern CFD simulators are themselves perfectly well-suited to act as (costly) ``layers'' in a deep network. Using well-studied adjoint methods, modern solvers can compute gradients of the output quantities of a simulation with respect to the input mesh. This allows us to integrate a fast CFD simulation (made fast because it is operating on a much smaller mesh) into the network itself, and allows us to \emph{jointly} train the GCN and the mesh input into the simulation engine, all in an end-to-end fashion.

We demonstrate that this combined approach performs substantially better than the coarse CFD simulation alone (i.e., the network is able to provide higher fidelity results than simply running a faster simulation to begin with), \emph{and} generalizes to novel situations much better than a pure graph-network-based approach. Moreover, the approach is still substantially faster than running the CFD simulation on the original size mesh itself. We believe that in total this represents a substantial advance towards integrating deep learning and existing state-of-the-art simulation software.

\begin{figure*}[t]
\vskip 0.2in
\begin{center}
\centerline{\includegraphics[width=\linewidth]{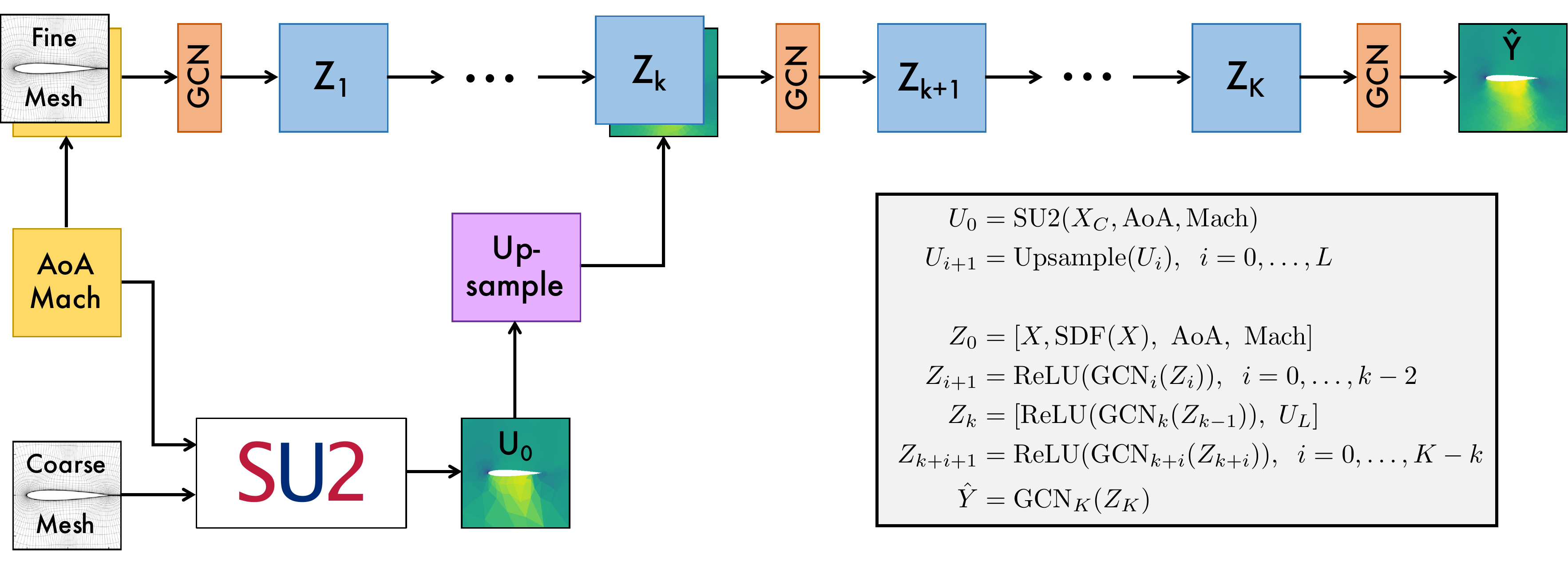}}
\caption{A diagram of the CFD-GCN model and its corresponding equations.}
\label{fig:diagram}
\end{center}
\vskip -0.2in
\end{figure*}

\bigskip
\section{Related Work}
\label{related}

This paper relates to previous work in many different areas at the interface of machine learning and fluid dynamics. We explore these diverse connections below.

\paragraph{Machine learning and CFD.} 
Many recent works have explored the interface between machine learning and CFD.
In many cases, the approach has been model-free, aiming to directly learn to predict physical processes using solely deep learning methods \citep{afshar_prediction_2019, guo2016convolutional}.
In work similar to the one presented here, \citet{afshar_prediction_2019} applied an encoder-decoder convolutional architecture to the task of predicting flow fields around an airfoil.
This paper, however, worked on fields represented as small images, instead of actual meshes. 
Moreover, generalization was not demonstrated at a range of parameters that would generate behavior significantly different from the ones seen during training.

Others have looked at employing deep learning models as function approximators that substitute certain terms in equations of interest, such as the turbulence terms in turbulence modeling \citep{duraisamy_turbulence_2019, king_deep_2018, singh_augmentation_2017}.
In contrast to these approaches, which embed deep learning models as a component in larger physical models, our approach aims to embed a full physical simulation into a deep learning system. 

\paragraph{Machine learning and graphics.} 
Though our method relates more closely to CFD, many papers have also explored applications of machine learning to fluid simulations for graphics. 
Unlike in CFD, fluid animations have the main goal of looking realistic, not necessarily aiming to model physical laws or conform perfectly to reality.
Graphics applications are frequently more naturally suited to deep learning methods, as in many such applications the simulations are natively performed in structured grids.
Some success has been achieved in generating realistic animations of smoke or water \citep{wiewel_latent-space_2018, kim_deep_2018, um_liquid_2017}.

Additionally, machine learning methods have also been applied to particle-based methods \citep{macklin_position_2013} in order to develop a differentiable fluid simulator \citep{schenck_spnets:_2018}.
Since these methods do not focus on accurately modeling the physical processes, having as their main goal generating realistic animations, they are unsuited for CFD tasks such as predicting aerodynamic flows for practical applications, which we explore in this work.

\paragraph{Adjoint-based differentiation.}
A central point of this work is the usage of a \emph{differentiable} fluid dynamics simulator, which allows us to embed a fluid simulation as a module in a deep learning model. 
Differentiation of PDE solvers has been used for decades for shape optimization in aerodynamics~\cite{jameson1988}, with diverse formulations, such as the continuous adjoint~\cite{economon2015a} and the discrete adjoint~\cite{albring2015,albring2016} being available.
In this work, we utilize SU2~\citep{economon2016}, which is an open-source suite that provides both CFD simulations and adjoint-based differentiation. 

The adjoint method has also been applied in the graphics literature, but with different goals.
\citet{mcnamara_fluid_2004} developed a differentiable fluid simulator for controlling smoke animations by optimizing forces acting on the smoke in order to have it match target shapes.

\paragraph{Graph neural networks.}
As a consequence of working with unstructured meshes, our proposed method makes use of many recent advances in graph neural networks.
The graph convolution operation we use in our models was originally proposed by \citet{kipf2016semi}, though many other approaches have also been proposed \citep{bronstein2017geometric, defferrard2016convolutional, hamilton2017inductive}.

Others have worked on applying graph neural networks to meshes \citep{hanocka2019meshcnn} or on graphs with positional information \citep{qi2017pointnet++}. However, in their work, the modifications to the mesh do not attempt to preserve or improve its functionality, serving only the purpose of pooling for a classification task. 
\citet{alet_graph_2019} employed graph neural networks with positional information to mesh a continuous space and model spatial processes. In this work, the dynamics were learned solely by the graph neural network, without the usage of any PDE solver.
To the best of our knowledge, our work is the first to directly modify a mesh to optimize its functionality for a downstream task, through using it on a differentiable simulator.
\bigskip

\section{Methodology}
\label{methodology}

Here we describe the general outline of our hybrid CFD simulation and graph neural network approach.  Based on this hybrid nature, we refer to our model as CFD-GCN. We first describe its broad architecture and then its different components in detail.  Finally, we describe the procedures used to train the network itself.

\subsection{The CFD-GCN Architecture}

The overall architecture of the CFD-GCN is shown in Figure \ref{fig:diagram}. Intuitively, the network operates over two different graphs, a ``fine'' mesh over which to compute the CFD simulation, and a ``coarse'' mesh (initially a simple coarsened version of the fine mesh, but eventually tuned by our model) that acts as input to the CFD solver.  As input, the network takes a small number of parameters that govern the simulation. For the case of the experiments in this work, in which we predict the flow fields around an airfoil, these parameters are the Angle of Attack (AoA) and the Mach number. These parameters are provided to the CFD simulation and are also appended to the initial GCN node features. Although this may seem to be a relatively low-dimensional task, even these two components can vary the output of the simulation drastically and are difficult for traditional models to learn when generalizing outside the precise range of values used to ``train'' the network.
The network operates by first running a CFD simulation on the coarse mesh, while simultaneously processing the graph defined by the fine mesh with GCNs. It then upsamples the results of the simulation, and concatenates these with an intermediate output from a GCN.  Finally, it applies additional GCN layers to these joint features, ultimately predicting the desired output values (in this case, the velocity and pressure fields at each node in the fine mesh).  We now describe each of these components in detail.

\paragraph{Graph structure and network input.}
The graph structure we use for the CFD-GCN is directly derived from the mesh structure used by traditional CFD software to simulate the physical system.  Specifically, we consider a two-dimensional, triangular mesh $M = (X, E, B)$. The first element, $X \in \mathbb{R}^{N \times 2}$, is a matrix containing the (x, y) coordinates of the $N$ nodes that compose the mesh. The second, 
\begin{align*}
    E=\{(i_1, j_1, k_1), \dots, (i_M, j_M, k_M)\},
    \end{align*}
is a set of $M$ triangular elements defined by the indices $(i, j, k)$ of their component nodes. The third, 
\begin{align*}
    B = \{(i_1, b_1), \dots, (i_L, b_L)\},
\end{align*}
is a set of $L$ boundary points, defined as a pair consisting of the index of the node and a tag $b$ that identifies which boundary the point belongs to (\textit{e.g.} airfoil, farfield, etc.). 

Such a mesh $M$ clearly defines a graph $G_M = (X, E_G)$ whose nodes are the same $X$, and whose edges $E_G$ can be directly inferred from the mesh elements $E$. Conversely, a graph can also be converted into a mesh if the structure of its edges is appropriate and a set of boundary points $B$ is provided.

In addition to the fine mesh used to compute the CFD simulation, we also consider a coarse mesh, denoted $M_C$.  This mesh has the same structure as the fine mesh $M$, with the number of nodes downsampled by almost 20x, which thus allows for much faster simulation.  Although this mesh also technically defines a graph, we do not directly compute any GCN over this graph, but instead only use it as input to the simulation engine.

In addition to the graphs themselves, the model also receives as input two physical parameters that define the behavior of the flow around an airfoil: the angle of attack (AoA) and the Mach number.  These two parameters are both fed into the simulation and appended as initial node features for every node in the GCN.  These two parameters ultimately are the quantities that vary from simulation-to-simulation, and thus the main task of the GCN is to learn how to predict the resulting flow field from these two parameters that define the simulation. Even thought this input space is low dimensional, it still defines a complex task, since varying these parameters gives rise to diverse behaviors of the fluid flows, as demonstrated, for example, in our generalization experiment (Section~\ref{sec:gen}).

\begin{figure}[t]
\vskip 0.2in
\begin{center}
\centerline{\includegraphics[width=0.75\columnwidth]{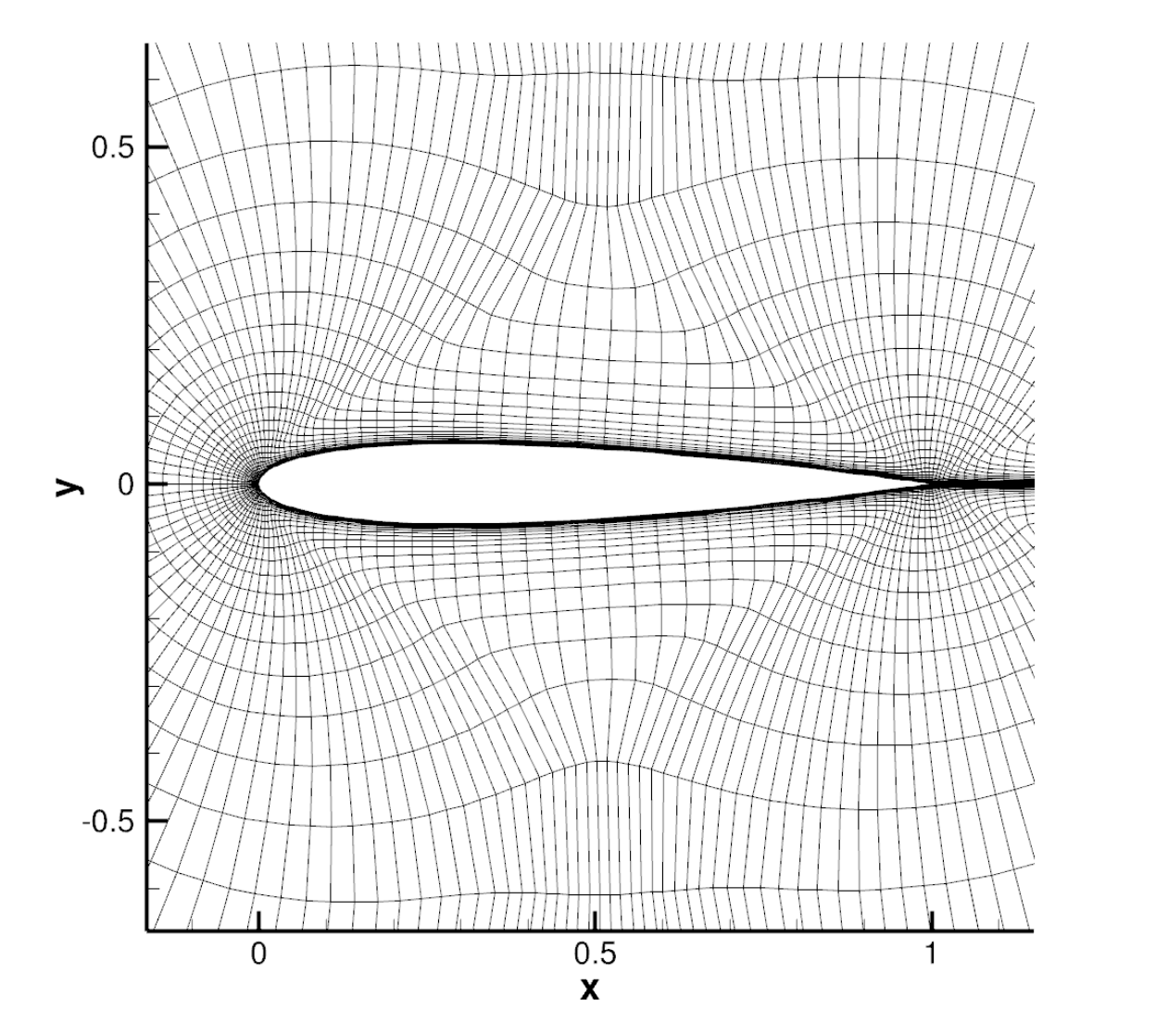}}
\caption{A NACA0012 mesh, zoomed in on the airfoil region.}
\label{fig:naca_mesh}
\end{center}
\vskip -0.2in
\end{figure}

\paragraph{The SU2 Fluid Simulator.}  A central component of the CFD-CGN model is the integrated differentiable fluid dynamics simulator.  As input, the fluid simulator takes coarse mesh $M_C$, plus the angle of attack and Mach number, and outputs predictions of the velocity and pressure at each node in the coarse graph.  We specifically employ the SU2 fluid simulator \citep{economon_su2:_2015}, an open source, industry-grade CFD simulation widely used by many researchers in aerospace and beyond.  Briefly, SU2 uses a finite volume method (FVM) to solve the Navier-Stokes equations over its input mesh.  Crucially for our purposes, the SU2 solver also support an adjoint method which lets us differentiate the outputs of the simulation with respect to its inputs and parameters (in this case, the coarse mesh $M_C$ itself, plus the angle of attack and Mach number).

Intuitively, the SU2 solver should be thought of as an additional layer in our network, which takes the angle of attack and Mach number as input, and produces the output velocity and pressure fields.  The equivalent of the ``parameters'' of a traditional layer is the coarse mesh itself: different configurations for the coarse mesh will be differently suited to integration within the remainder of the CFD-GCN.  Thus, the main learning task for the SU2 portion of our model is to \emph{adjust} the coarse mesh in a manner than eventually maximizes accuracy of the resulting full CFD-GCN model.  The adjoint method in SU2 uses reverse-mode differentiation, so gradients can be efficiently computed with respect to a scalar-valued loss such as the overall predictive error of the CFD-GCN.

Finally, although not strictly a research contribution, we want to mention that as part of this project we have developed an interface layer between the SU2 solver and the PyTorch library.  This interface allows full SU2 simulations to be treated just as any other layer within a PyTorch module, and we hope it will find additional applications at the intersection of deep learning and (industrial-grade) CFD simulation.  The code for the work presented in this paper can be found at \url{https://github.com/locuslab/cfd-gcn}.

\paragraph{Upsampling.}
The output of the coarse simulation described above is a mesh with the predicted values for each field at every node. For this to be used towards generating the final prediction, we need to upsample it to the size of the fine mesh.
We do this by performing successive applications of squared distance-weighted, k-nearest neighbors interpolation \citep{qi2017pointnet++}.

Let us call $U \in \mathbb{R}^{N_U \times 3}$ the upsampled version of some coarser graph $D \in \mathbb{R}^{N_D \times 3}$. For every row $U^{(i)}$, with corresponding node position $X_U^{(i)}$, we find the set $\{n_1, \dots, n_k\}$ containing the indices of the $k$ closest nodes to $X_U^{(i)}$ in the coarser graph $X_D$. Then, we define $U^{(i)}$ as
\begin{align*}
\label{eq:upsample}
    U^{(i)} = \frac{\sum_{j=1}^k w(n_j) D^{(n_j)}}{\sum_{j=1}^k w(n_j)},
\end{align*}
where 
\begin{align*}
    w(c_j) = \frac{1}{\|X_U^{(i)} - X_D^{(n_j)}\|_2^2}.
\end{align*}
As a default, we set $k=3$. 

\paragraph{Graph Convolutions.}
As depicted on Figure \ref{fig:diagram}, the output of the coarse simulation is processed by a sequence of convolutional layers.
In order to operate directly on the mesh output of the CFD simulation, we utilize the graph convolutional network (GCN) architecture from \citet{kipf2016semi}. 
This architecture defines a convolutional layer for graphs.

A general graph consisting of $N_Z$ nodes, each with $F$ features, is defined by its feature matrix $Z_i \in \mathbb{R}^{N_Z \times F}$ and its adjacency matrix $A \in \mathbb{R}^{N_Z \times N_Z}$. We can then further define $\Tilde{B} = \Tilde{D}^{-\frac{1}{2}} (A + I) \Tilde{D}^{-\frac{1}{2}}$, where $I$ is the identity matrix and $\Tilde{D}$ the diagonal degree matrix, with its diagonal given by $\Tilde{D}_{ii} = 1 + \sum^{N_Z}_{j=0} A_{ij}$. Then, a GCN layer with $F$ input channels and $F'$ output channels, parameterized by the weight matrix $W \in \mathbb{R}^{F \times F'}$ and the bias term $b \in \mathbb{R}^{N_Z \times F'}$, followed by a ReLU non-linearity, will have as output
\begin{align*}
    \tilde{Z}_{i+1} &= \tilde{B} Z_i W_i + b_i \equiv \mathrm{GCN}_i(Z_i). \\
    Z_{i+1} &= \text{ReLU}(\tilde{Z}_{i+1})
\end{align*}

\paragraph{CFD-GCN.} With all the components of the CFD-GCN desribed above, we can now bring them all together to describe the full pipeline depicted in Figure~\ref{fig:diagram}.

First, an SU2 simulation is run with the coarse mesh and the physical parameters. The output of this coarse simulation is upsampled $L$ times. 
\begin{equation}
\begin{aligned}
\label{eq:up}
    U_0 & = \text{SU2}(X_C, \text{AoA}, \text{Mach}) \\
    U_{i+1} & = \text{Upsample}(U_i), \;\; i=0,\ldots,L.
\end{aligned}
\end{equation}

Concurrently, the fine mesh has the physical parameters and the signed distance function (SDF) appended to each of its nodes' features. 
The resulting graph is then passed through a series of graph convolutions. At some specified convolutional layer $k$, the final upsampled value $U_L$ is appended to the output $Z_k$ of the $k$-th convolution. Another set of convolutions is performed in order to generate the final prediction $\hat{Y}$
\begin{align}
\label{eq:gcn}
    Z_0 & = [X, \text{SDF}(X),\ \text{AoA},\ \text{Mach}] \nonumber \\
    Z_{i+1} & = \text{ReLU}(\text{GCN}_i(Z_i)), \;\; i=0,\ldots,k-2 \nonumber \\
    Z_k & = [\text{ReLU}(\text{GCN}_k(Z_{k-1})),\ U_L] \\
    Z_{k+i+1} & = \text{ReLU}(\text{GCN}_{k+i}(Z_{k+i})), \;\; i=0,\ldots,K-k \nonumber \\
    \hat{Y} & = \text{GCN}_{K}(Z_K). \nonumber 
\end{align}
 Here, $[\cdot, \cdot]$ is the matrix concatenation operation over the column dimension.

\begin{figure}[t!]
    \vskip 0.7in
    \begin{center}
    \centerline{\includegraphics[width=0.5\columnwidth]{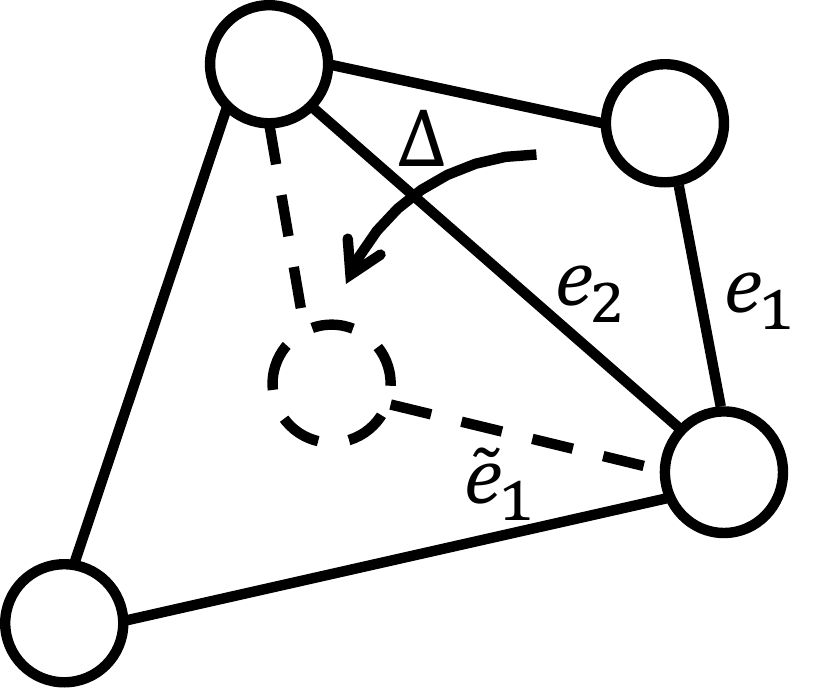}}
    \end{center}
    \vskip 0.05in
    \caption{For certain updates to the mesh ($\Delta$), a node might be pushed over the edge of its triangular element, generating overlap of elements. These non-physical situations harm convergence of the simulation. When this happens, the cross product between ordered edges changes. Before the update, $e_1 \times e_0 > 0$, while afterwards $\tilde{e}_1 \times e_2 < 0$.}
    \label{fig:flip}
\end{figure}

\subsection{Training the CFD-GCN}
Given that the entire CFD-GCN as formulated above can be treated as a single differentiable deep network (including the SU2 ``layer'' discussed above), the training process itself is largely straightforward. The model is trained to predict the output fields $Y \in \mathbb{R}^{N \times 3}$, consisting of the x and y components of the velocity and the pressure at each node in the fine mesh, by minimizing the mean squared error (MSE) loss $\ell$ between the prediction $\hat{Y}$ and ground truth
\begin{align*}
\label{eq:loss}
    \ell(Y, \hat{Y}) = \frac{1}{3N}\|Y - \hat{Y}\|_2^2,
\end{align*}
where the ground truth $Y$ in this case is obtained by running the full SU2 solver to convergence on the original fine mesh.  

The training procedure optimizes the weight matrices $W_i$ and $b_i$ of the GCNs, and the positions of the nodes in the coarse mesh $X_C$ by backpropagating through the CFD simulation.
The loss is minimized using the Adam optimizer \citep{kingma2014adam} with a learning rate $\alpha = 5 \cdot 10^{-5}$.

\begin{figure}[t!]
\vskip 0.2in
\begin{center}
\centerline{\includegraphics[width=\columnwidth]{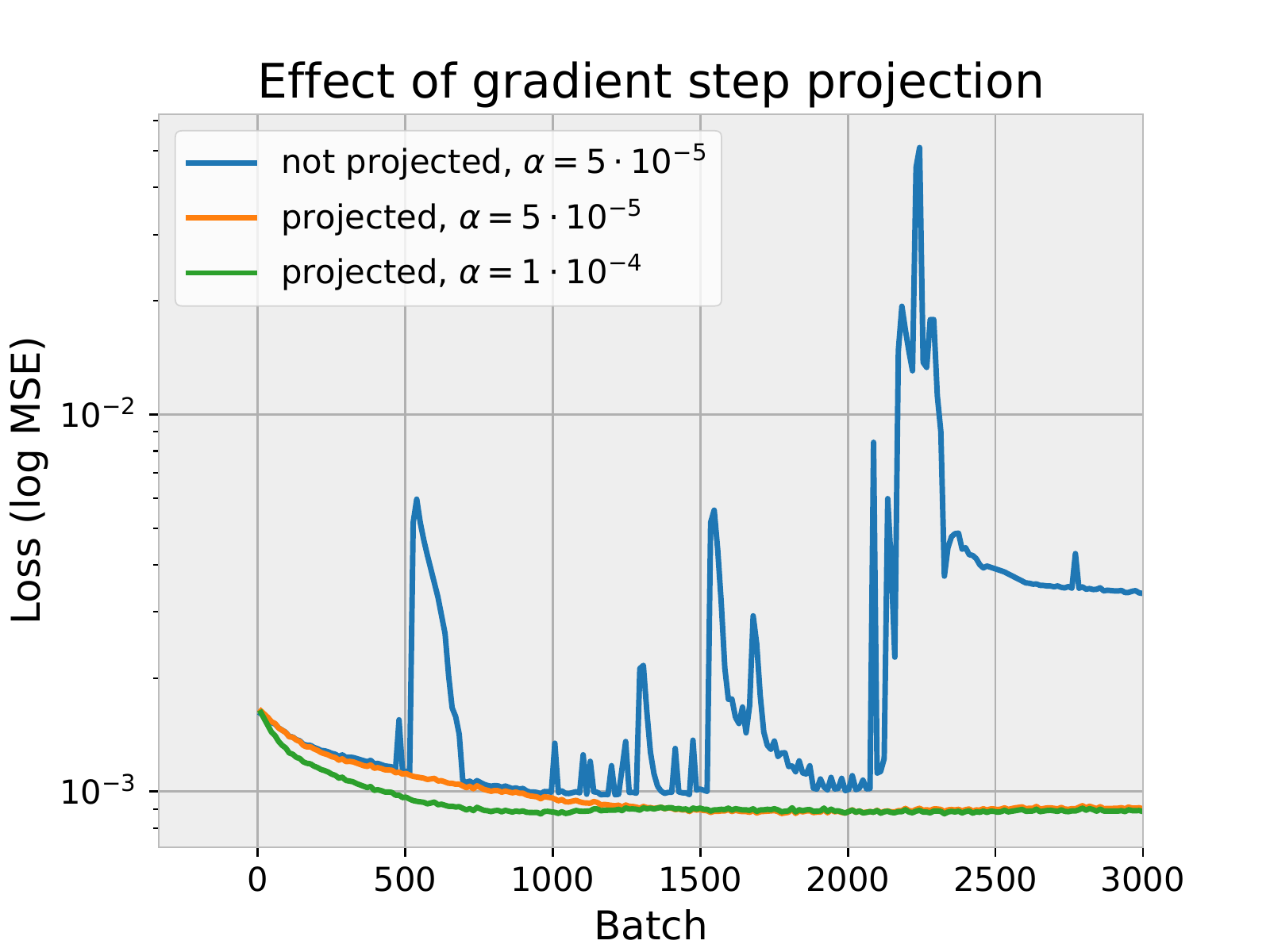}}
\caption{Optimizing a mesh without correcting the gradient update to prevent degeneration causes the training to diverge. Meshes trained using the projected gradient step (avoiding non-physical elements) learn smoothly, even with a higher learning rate $\alpha$.}
\label{fig:flip_curve}
\end{center}
\vskip -0.2in
\end{figure}

\paragraph{Mesh degeneration.} An issue arises when optimizing the input coarse mesh. Gradually, as the node positions are moved by the gradient descent updates, it is possible that, in a given triangular element, one of its nodes crosses over an edge (see Figure~\ref{fig:flip}). This generates non-physical volumes, which harm the stability of the simulations, frequently impeding convergence. In other words, at each gradient update step, our optimizer updates the mesh nodes by performing the update
\begin{align*}
    X_C \leftarrow X_C + \Delta X_C,
\end{align*}
with some small update matrix $\Delta X_C$ of the same shape as $X_C$. If left unmodified, this $\Delta X_C$ can cause the aforementioned issue. 

In order to avoid this, we seek to generate a projected update $P(\Delta X_C)$ such that only non-degenerating updates are performed. We start with $P(\Delta X_C) = \Delta X_C$. Then, we check which elements in the mesh have a node pushed over an edge by $\Delta X_C$. This can be done by computing the cross product of two edges in each triangular element in a consistent order. If the sign of the cross product flips with the update $X_C + \Delta X_C$, that means a node crossed over an edge (since this causes the ordering of the nodes to change). 
This is depicted in Figure~\ref{fig:flip}, where the cross product of the edges $e_1$ and $e_2$ is positive before the update, but negative afterwards.  

For every element $E=(i, j, k)$ which has flipped, we set the rows $i$, $j$ and $k$ of $P(\Delta X_C)$ to $0$, thus performing no updates to those points in $X_C$. Since removing the updates to some nodes might cause new elements to flip, this procedure is repeated until no points are flipped. Once we reach this state, we perform the projected gradient update
\begin{align*}
    X_C \leftarrow X_C + P(\Delta X_C).
\end{align*}

In Figure~\ref{fig:flip_curve} we see the results of optimizing the nodes of a mesh to improve a prediction loss both with and without the correction to the gradient update. Whereas the mesh optimized without the correction quickly degenerates and the loss diverges, the one with the projected gradient update learns smoothly, even for a higher learning rate $\alpha$.

\begin{figure}[t!]
\vskip 0.2in
\begin{center}
\centerline{\includegraphics[width=\columnwidth]{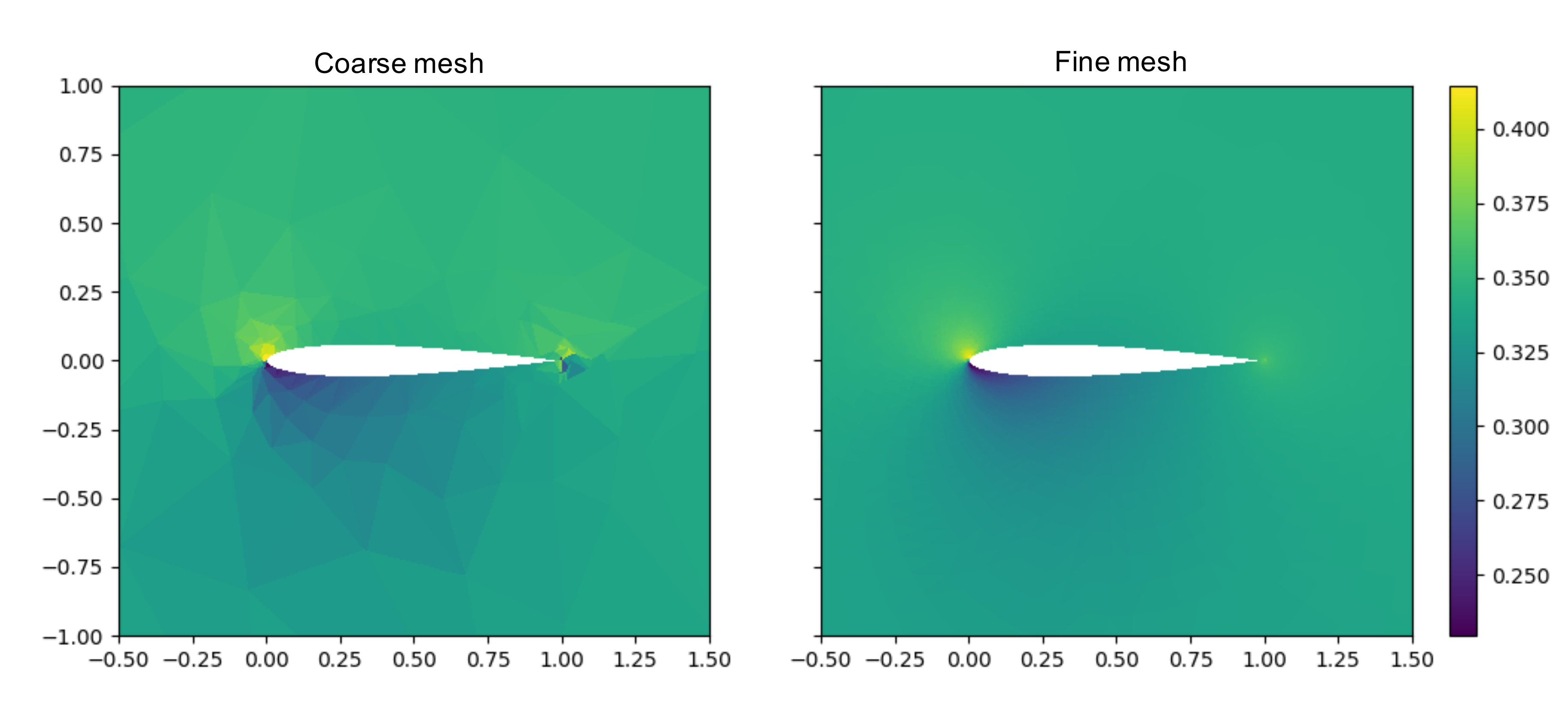}}
\caption{Two simulations with identical physical parameters, one with the coarse mesh (left) and one with the fine mesh (right). The pressure component of the output field is presented here. The coarse elements are easily noticeable on the left, whereas on the fine mesh on the right the elements are small enough to be barely visible at this size.}
\label{fig:sims}
\end{center}
\vskip -0.2in
\end{figure}

\section{Experiments}
\label{experiments}

For all experiments we use the NACA0012 airfoil, represented as a fine mesh with 6648 nodes Figure~\ref{fig:naca_mesh}). The coarse mesh for the same airfoil has 354 nodes. Both meshes are mixed triangular and quadrilateral meshes, but for usage with the CFD-GCN model the coarse mesh is converted to purely triangular by dividing every quadrilateral element in half along a diagonal. 
All meshes were created using Pointwise Mesh Generation Software\footnote{https://www.pointwise.com}.

All CFD simulations are performed by solving the0.0 steady-state, compressible, inviscid case of the Navier-Stokes equations (the Euler equations) using SU2. Ground truth simulations on the fine mesh are run to convergence, while simulations on the coarse mesh are run for up to 200 iterations. 
Sample outputs of simulations with identical physical parameters in each mesh are presented in Figure~\ref{fig:sims}.

For the CFD-GCN model, we set $k=3$, $K=6$, and $L=1$. That is, we perform one upsampling step on the coarse simulation, appending it to the third GCN layer, and then perform 3 additional convolutions, for a total of 6 GCN layers. All GCNs are set to have 512 hidden channels. A batch size of 16 is used on all experiments.

In the experiments, our model is compared to three baselines that can be interpreted as ablated versions of the full CFD-GCN model: the ``upsampled coarse mesh'' (UCM) baseline, a pure GCN baseline, and the ``frozen mesh'' version of the CFD-GCN. Each of these demonstrates the importance of each part of the full proposed model.
The upsampled coarse mesh baseline consists simply of the part of the model described in Equations~\ref{eq:up}. That is, simply of running the simulation on the coarse mesh and interpolating the output up to the full mesh size. It does not have convolutional layers and it does not perform any learning. 
The GCN baseline, conversely, consists solely of the GCNs, without the simulation. That is, the part of the model described in Equations~\ref{eq:gcn} (without the appended $U_L$). The GCNs are set to the same parameters as used for the CFD-GCN (6 layers with 512 hidden channels).
Finally, the ``frozen mesh'' version of the CFD-GCN consists of the full CFD-GCN model, with both the GCNs and the coarse simulation, but the gradients through the fluid simulation are not computed, and thus the coarse mesh is not optimized (it is therefore ``frozen'' through training).

\begin{figure}[t!]
\vskip 0.2in
\begin{center}
\centerline{\includegraphics[width=\linewidth]{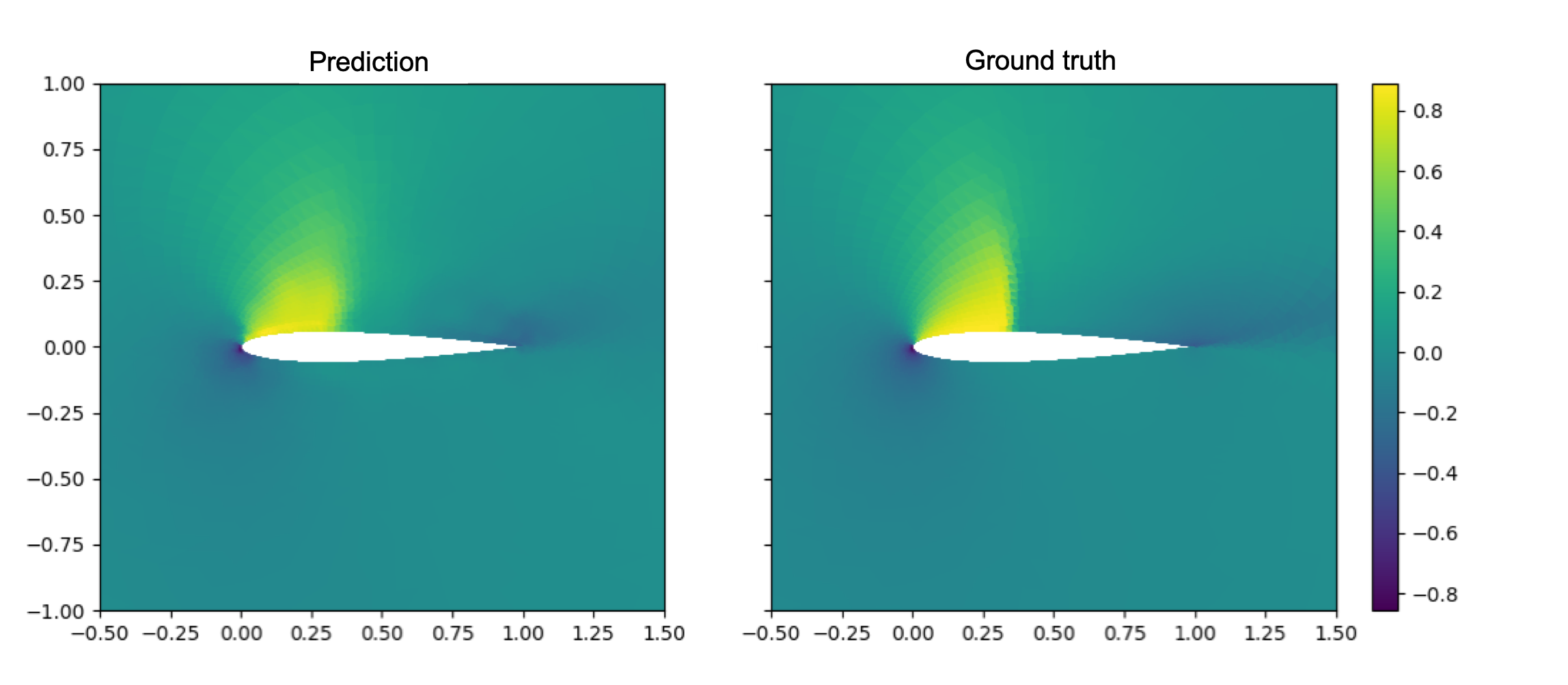}}
\caption{CFD-GCN model prediction and ground truth for a test sample in the interpolation task. The x component of the velocity output field is presented here. The other fields can be found in Figure~\ref{fig:pred_rest} in the Appendix}
\label{fig:pred}
\end{center}
\vskip -0.2in
\end{figure}

\subsection{Interpolation}
\label{sec:pred}

In order to test our model's ability to make accurate flow field predictions, we test its predictions across a range of different physical parameters. We construct training and test sets composed of values for the AoA and Mach number. 

The training set is defined by
\begin{align*}
    \text{AoA}_\text{train} &= \{-10, -9, \dots, 9, 10\}, \\
    \text{Mach}_\text{train} &=  \{0.2, 0.3, 0.35, 0.4, \\
    \ & \qquad  0.5, 0.55, 0.6, 0.7\}.
\end{align*}
\medskip

Similarly, test pairs are sampled from the sets
\begin{align*}
    \text{AoA}_\text{test} &= \{-10, -9, \dots, 9, 10\}, \\
    \text{Mach}_\text{test} &=  \{0.25, 0.45, 0.65\}.
\end{align*}
Training pairs are then sampled uniformly from $\text{AoA}_\text{train} \times \text{Mach}_\text{train}$, and test pairs from $\text{AoA}_\text{test} \times \text{Mach}_\text{test}$.
Here we can see that even though the train and test set are different, the parameters come from similar ranges, and the two sets contain examples with a similar range of qualitative behaviors. This experiment therefore tests the ability of our model to interpolate from parameters seen in training to unseen, yet similar ones at test time. 
Even though this procedure does not present a strong test of the learning ability of the model, it is a common form of evaluation in many works that apply deep learning methods to CFD (e.g., \citet{afshar_prediction_2019, guo2016convolutional}).
We present a stronger test of generalization to new scenarios in our next experiment (Section~\ref{sec:gen}).

The model takes in as input the pairs and, using the coarse mesh, predicts the three components of the output field, as described in Section~\ref{methodology}. These predictions are compared against ground truth simulations performed on the fine mesh.

\begin{figure}[t]
\vskip 0.2in
\begin{center}
\ifdefined\logplots
    \centerline{\includegraphics[width=\linewidth]{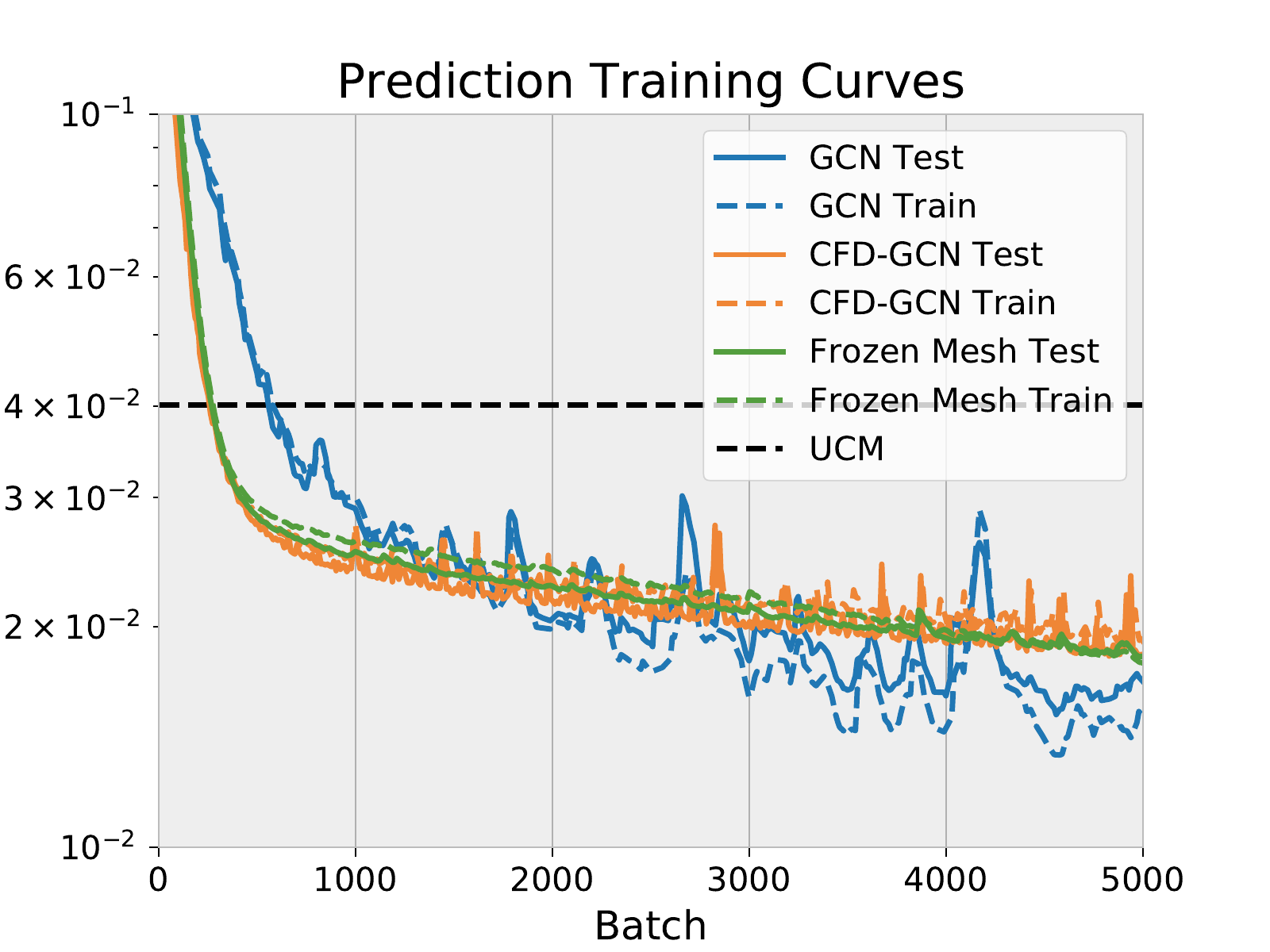}}
\else
    \centerline{\includegraphics[width=\linewidth]{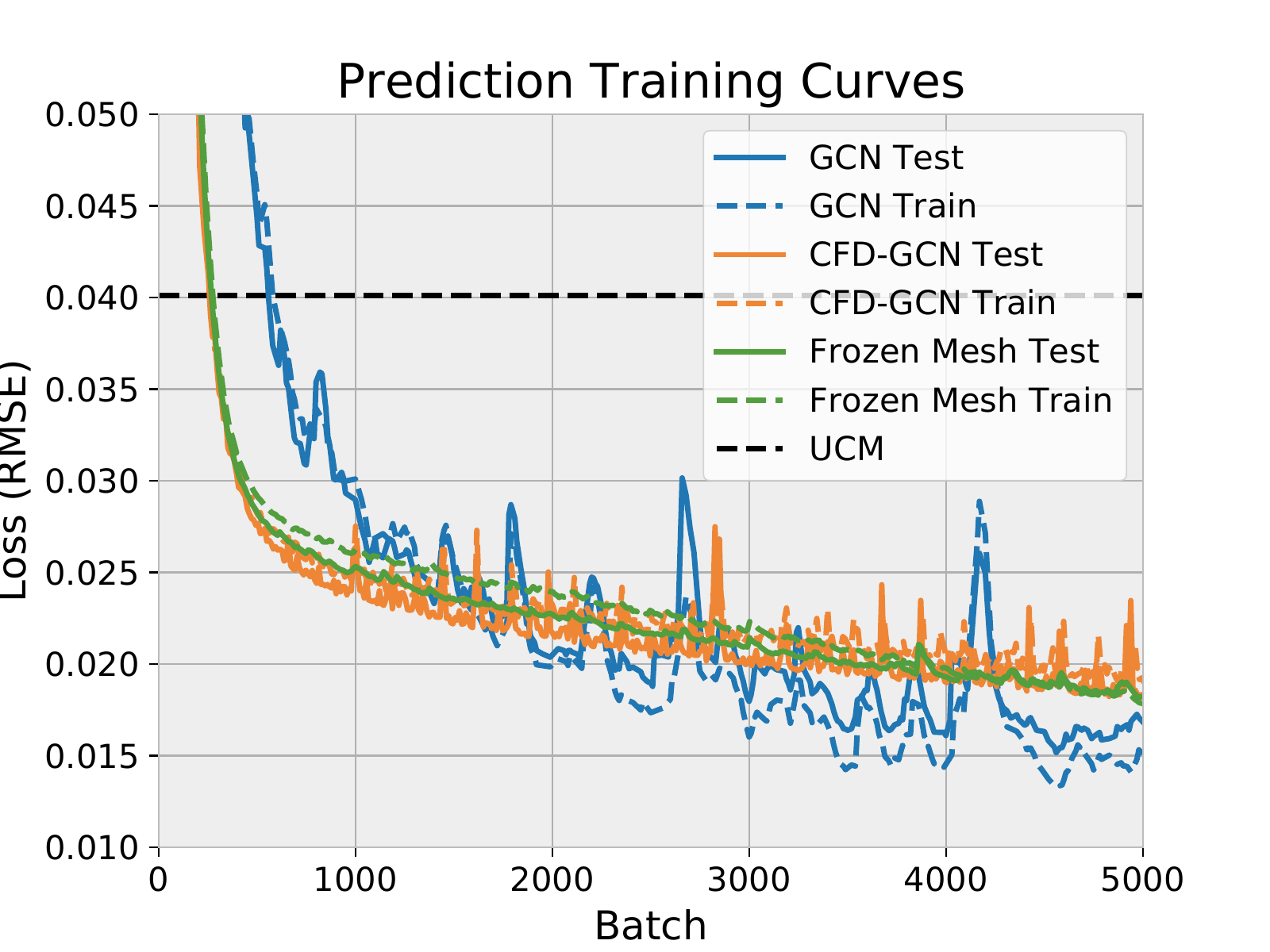}}
\fi
\caption{Training curves for the interpolation experiment. The vertical axis represented the root mean squared error (RMSE). The GCN baseline overfits more strongly to the training set, but since the test set is drawn from a similar distribution of parameters, this helps it outperform the CFD-GCN model.}
\label{fig:pred_curve}
\end{center}
\vskip -0.2in
\end{figure}

Results are summarized in Table~\ref{tab:res}. A sample prediction is presented in Figure~\ref{fig:pred}.
We can see from the results that our method outperforms the upsample coarse mesh baseline. This superiority to the upsample coarse mesh baseline demonstrates that the model is not simply upsampling the coarse prediction. The processing done by the GCNs is in fact improving its predictions.
We can also observe that the CFD-GCN performs worse than the GCN baseline.
The fact that the CFD-GCN underperforms the GCN baseline on the test set is a consequence of the similarity of the settings between training and testing, as we will see in the next experiment. The GCN is capable of overfitting the training set better (as we can see in Figure~\ref{fig:pred_curve}) therefore it also performs well on the very similar test set.

\subsection{Generalization}
\label{sec:gen}

\begin{figure}[t]
\vskip 0.2in
\begin{center}
\centerline{\includegraphics[width=\linewidth]{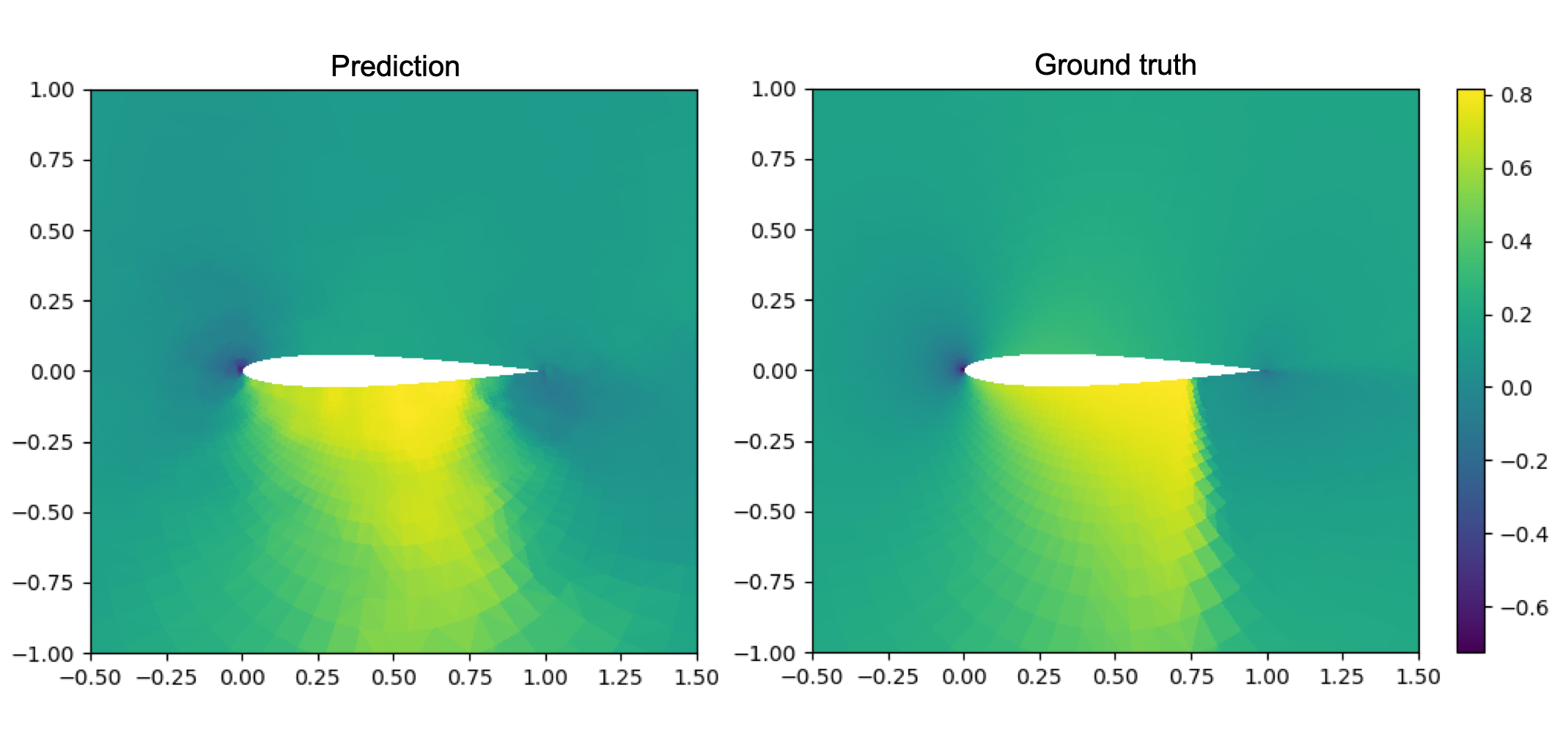}}
\caption{The CFD-GCN model prediction for a test sample with a large shock in the generalization task. It can generalize better than the pure GCN model to examples with large shocks, which were not seen in the training set. The x component of the velocity output field is presented here. The other fields can be found in Figure~\ref{fig:cfd_gen_rest} in the Appendix}
\label{fig:cfd_gen}
\end{center}
\vskip -0.2in
\end{figure}

Depending on the parameter configuration for a given simulation, a ``shock'' may or may not form around the airfoil. Figure~\ref{fig:cfd_gen} presents an example configuration in which we observe a shock. As can be noticed from the figure, these shocks present qualitatively different behavior from the smooth flow fields of ``regular'' simulations. In this experiment, we aimed to use such a difference in behavior in order to test the generalization capabilities of our model.

We thus constructed a training split such that there were no simulations with a shock present in the training set.
To achieve this goal, we used the same data points consisting of pairs of AoA and Mach parameters as in the last experiment. Here, however, the points were split into train and test set such that all points with a Mach number greater than 0.5 were placed in the test set.
Shocks become very frequent as the Mach number increases.
In order to ensure this qualitative split between training and test sets, the ground truth simulation for each pair of parameters was analyzed individually to guarantee no simulations with shocks put into the training set. 

This particular training split generates a very strong test of generalization.
Not only does the test set present behavior that is qualitatively different from what is observed in the training set, it also contains a significant quantitative difference, due to the wide range of Mach numbers that are never seen in training.
Therefore, this experiment presents a good setting to evaluate the generalization capabilities of the proposed model and the baselines. 

\begin{figure}[t]
\vskip 0.2in
\begin{center}
\centerline{\includegraphics[width=\linewidth]{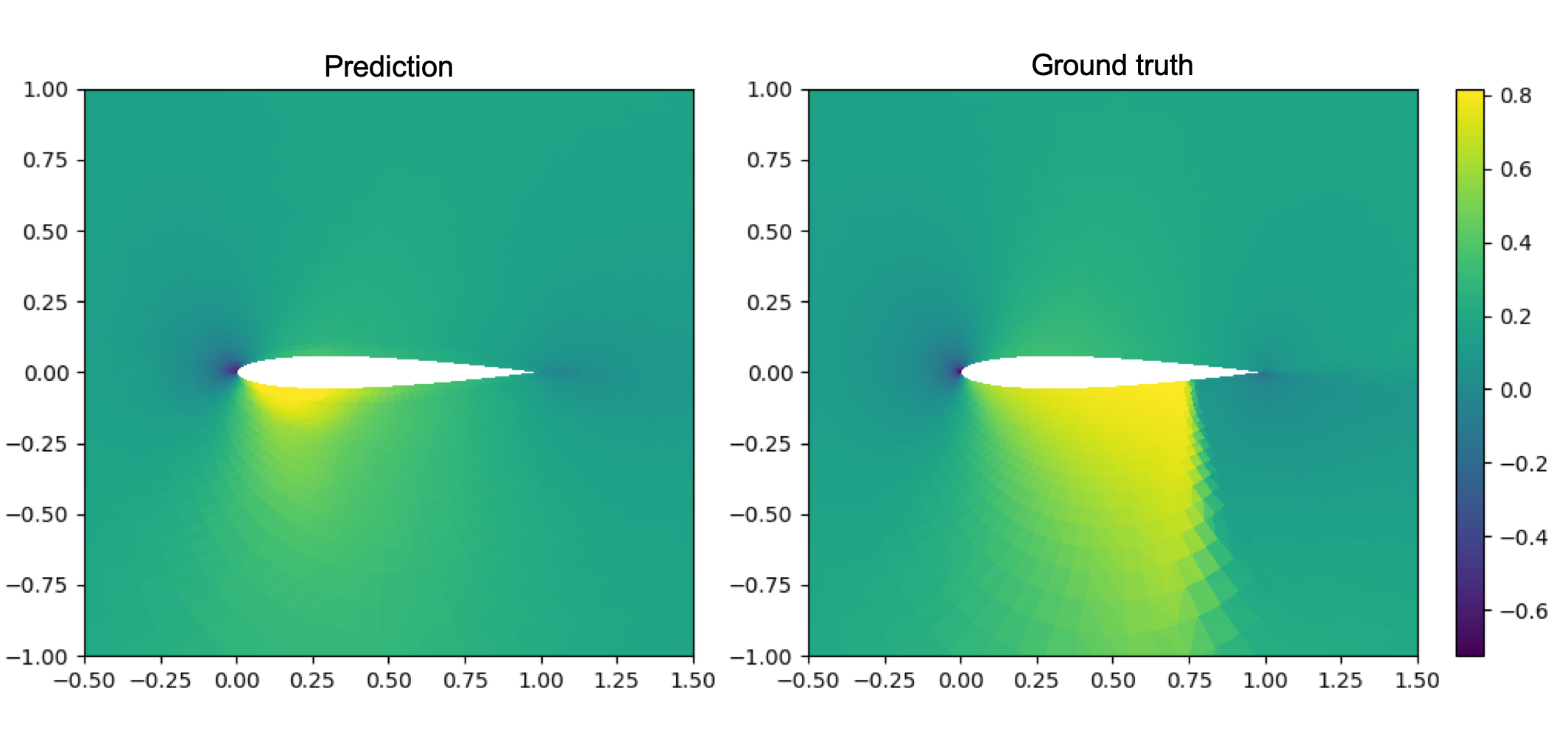}}
\caption{The GCN baseline prediction for a test sample with a large shock in the generalization task. In many cases with large shocks the GCN is unable to generalize to this previously unseen behavior. The x component of the velocity output field is presented here. The other fields can be found in Figure~\ref{fig:gcn_gen_rest} in the Appendix}
\label{fig:gcn_gen}
\end{center}
\vskip -0.2in
\end{figure}

\begin{table*}[t]
\caption{\textbf{Interpolation and generalization tasks. }Test root mean squared error (RMSE) for the interpolation and generalization tasks. The CFD-GCN model is compared to the frozen mesh, upsampled coarse mesh (UCM) and the pure GCN model baselines. The CFD-GCN and the GCN achieve similar performance in the interpolation task. The slightly better performance of the GCN is due to overfitting to the training distribution, as demonstrated by the superior performance of the CFD-GCN in the generalization task.
\textbf{Runtime.} Runtimes for a batch of 16 predictions compared to ground truth simulations with the fine mesh. The CFD-GCN runs significantly faster than running a full simulation, while presenting better results than the GCN. Results are for evaluation mode, without the backwards pass. Tests performed on a 24-core, 2.2 GHz machine with an NVidia GTX 2080 GPU.}
\label{tab:res}
\vskip 0.15in
\begin{center}
\begin{small}
\begin{sc}
\begin{tabular}{lcc|cr}
\toprule
Model & Interpolation (RMSE) & Generalization (RMSE) & Batch Prediction Time (s)  \\
\midrule
CFD-GCN & $1.8 \cdot 10^{-2}$ & $\mathbf{5.4 \cdot 10^{-2}}$ & 2.0 \\
Frozen Mesh & $1.8 \cdot 10^{-2}$ & $6.1 \cdot 10^{-2}$ & 2.0 \\      
Upsampled Coarse Mesh & $4.0 \cdot 10^{-2}$ & $7.0 \cdot 10^{-2}$ & 1.9 \\        
GCN & $\mathbf{1.4 \cdot 10^{-2}}$ & $9.5 \cdot 10^{-2}$ & 0.1 \\      
Ground Truth Simulation & -- & -- & 137 \\
\bottomrule
\end{tabular}
\end{sc}
\end{small}
\end{center}
\vskip -0.1in
\end{table*}

\begin{figure}[t]
\vskip 0.2in
\begin{center}
\ifdefined\logplots
    \centerline{\includegraphics[width=\linewidth]{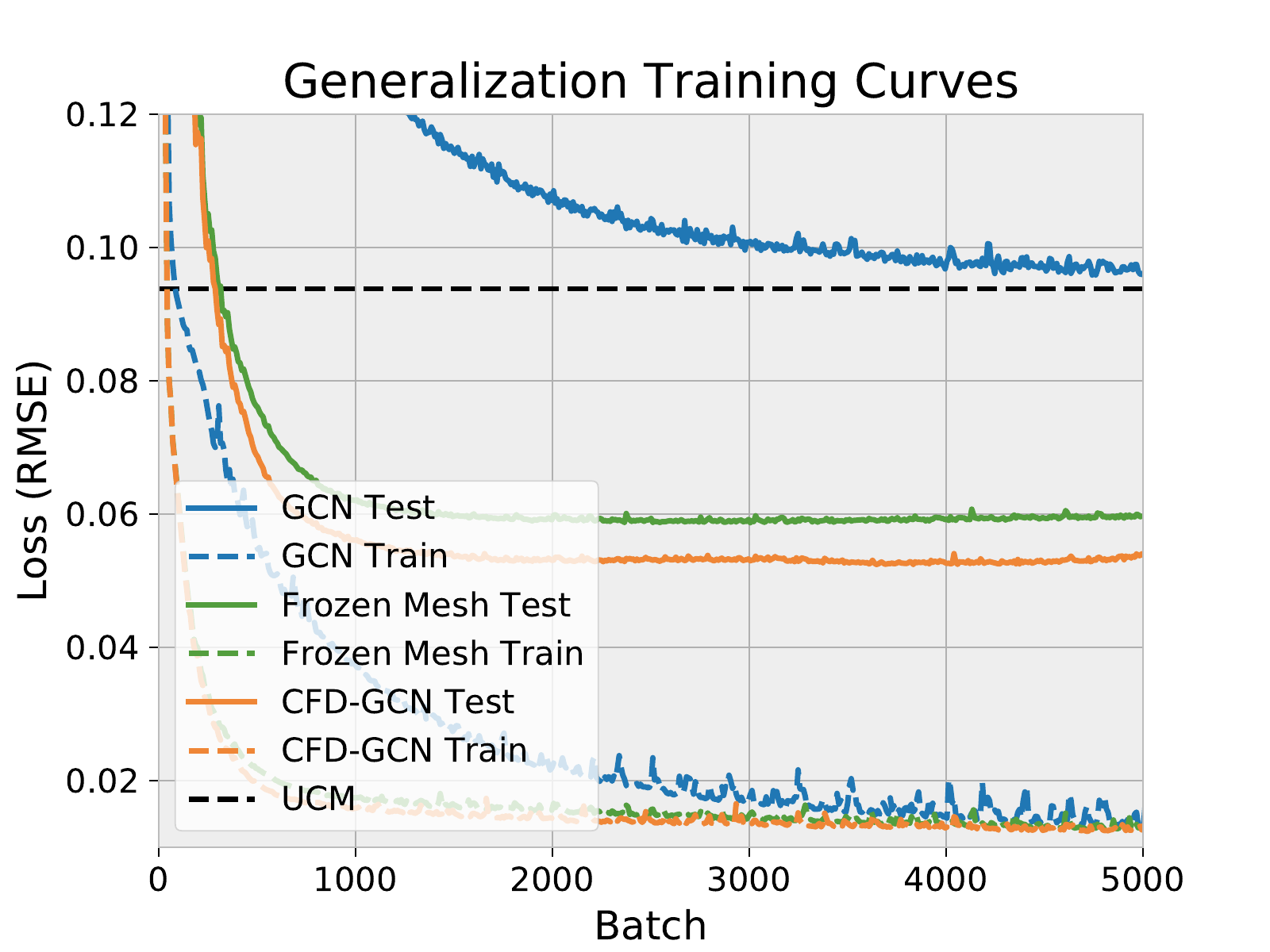}}
\else
    \centerline{\includegraphics[width=\linewidth]{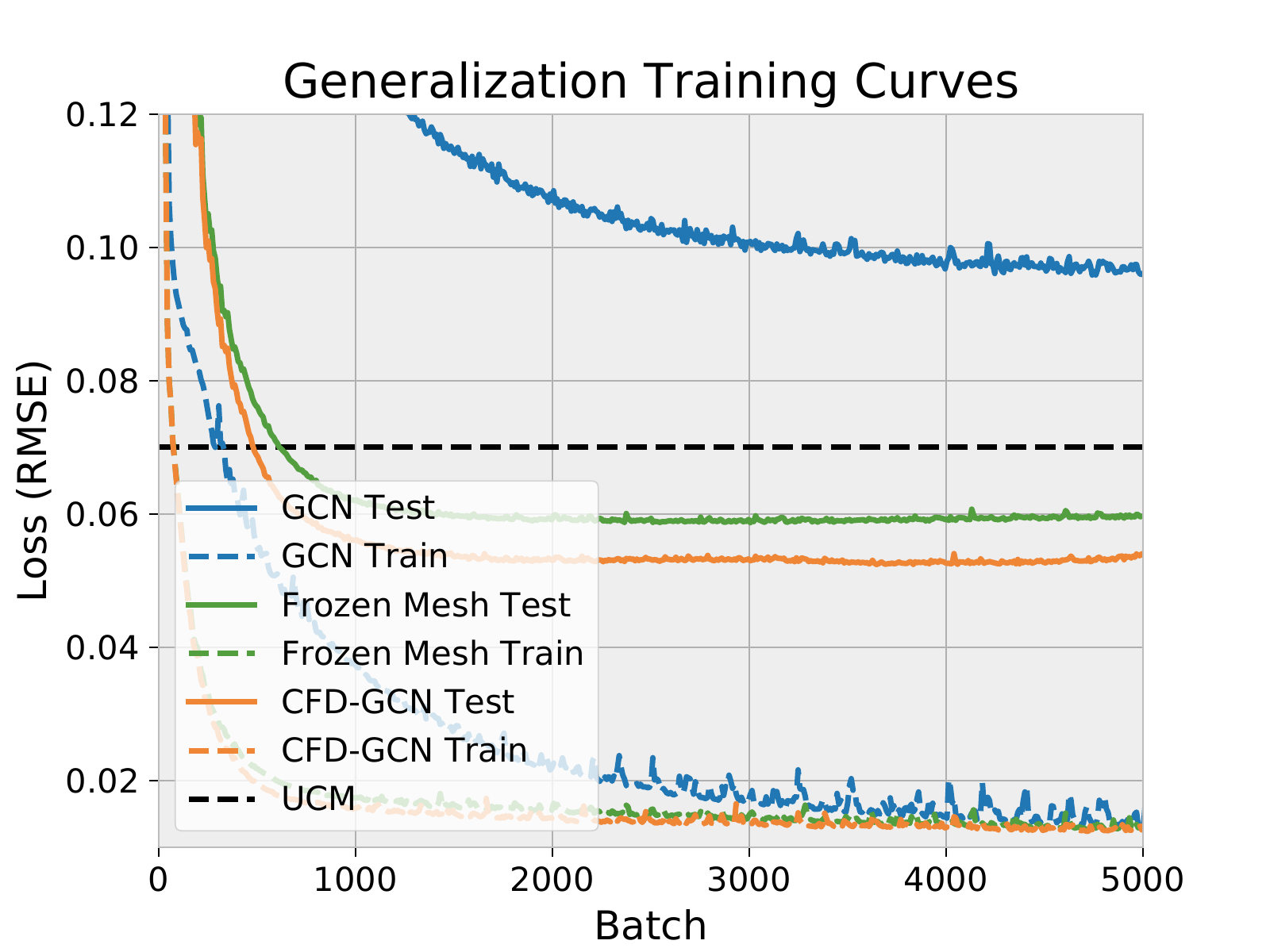}}
\fi
\caption{Training curves for the generalization experiment. The GCN baseline overfits more strongly to the training set, being unable to generalize well to the test set.}
\label{fig:gen_curve}
\end{center}
\vskip -0.2in
\end{figure}

Table~\ref{tab:res} summarizes the results for this experiment, and Figure~\ref{fig:gen_curve} presents the training curves for the CFD-GCN and the baselines. 
As expected, we can see that the CFD-GCN model generalizes better to the test set containing unseen shock behavior. 
This is also demonstrated qualitatively in Figure~\ref{fig:gcn_gen}, which presents a sample prediction from the GCN baseline. This baseline overfits the training set strongly, and is unable to consistently make predictions for simulations with shocks.
Conversely, even though it was never trained on this type of flow, our method is able to generate predictions that are closer to the ground truth by using the available coarse simulation. 
This can also be observed in Figure~\ref{fig:cfd_gen}.
In many test cases containing shock, the CFD-GCN is able to approximate the characteristics even of the unseen behavior.
Additionally, the performance of the upsampled coarse mesh baseline demonstrates that once again our method is not relying simply on upsampling the simulation, but is also learning additional information to improve its predictions. Sample predictions for the upsampled coarse mesh baseline are presented in Figure~\ref{fig:ucm_gen} (in the Appendix).

Finally, we can also observe that the full CFD-GCN model also outperforms the frozen mesh baseline. This result demonstrates the optimizing the coarse mesh by using the gradients computed through the simulations allows the model to optimize the simulation outputs in order to achieve predictions that generalize better.
Figure~\ref{fig:opt_mesh} (in the Appendix) demonstrates the transformation of the coarse mesh before and after the training procedure. The optimization performed is significant enough that the changes are easily perceptible visually. The changes are greater around the airfoil, where the gradient of the prediction loss is expected to be higher, demonstrating that the training procedure adjusts the mesh according to the training objective.

\subsection{Runtime}

Table~\ref{tab:res} demonstrates the efficiency of our method compared to running a full simulation. 
By downsampling the mesh down almost 20x to 354 nodes, our method is able to make a prediction much faster than running the full ground-truth simulations.

In our experiments, and as can be noticed with the comparison to the upsampled coarse mesh and GCN baselines, we observed that the bulk of the time that the GCN takes to make a prediction is consumed by the CFD simulation. On average, approximately 85\% of the time to make a batch of predictions was due to the CFD simulation, 10\% to upsampling the mesh and 5\% to processing the graph convolutions. Due to the additional complexity of performing the simulations, total training time for the pure GCN baseline was also faster. Whereas training the CFD-GCN took approximately 19 hours, training the GCN baseline took approximately 1.3 hours.

Even though the pure GCN baseline model is able to make predictions faster, it does not generalize as well across diverse physical behaviors, as demonstrated in our experiments. Therefore, we note that the CFD-GCN model, as its name suggests, provides a trade-off between the high cost and ability to generalize of a full CFD simulation, and the low cost and ability to generalize of GCN predictions.

\section{Conclusion}
\label{conclusion}

In this paper, we have presented a system that integrates a differentiable CFD simulator as a module into a larger deep learning system. 
This system is unique in that it works directly with unstructured meshes, by using graph operators. 
As demonstrated by our experiments, the combination of the fluid simulations performed in a coarser version of the original mesh and the learned parts of the model, creates a system that is able to generate predictions that are faster than a full simulation and more accurate than a purely learned model. 
Moreover, we show that such a model is able to generalize to parameter settings outside of the training distribution. 
We believe the system we present in this work contributes to a recent trend of combining structured knowledge and learning-based methods. When correctly applied, these types of combinations have the potential to benefit from the complementary strong points of each approach.

\bibliography{refs}
\bibliographystyle{icml2020}

\appendix
\onecolumn
\icmltitle{Appendix}

\section{Multi-Airfoil Experiment}
\label{sec:multi}

As an additional test of the generalization capabilities of our model, we performed an experiment in which we train the CFD-GCN on two airfoils and test on a previously unseen one. For this task, we use the same range of physical parameters as before on all airfoils. The training set is composed of simulations using the NACA4412 and the RAE2822 airfoils. The training set uses the same NACA0012 airfoil as the previous tasks.

\begin{figure}[h!]
\vskip 0.2in
\begin{center}
\ifdefined\logplots
    \centerline{\includegraphics[width=0.5\linewidth]{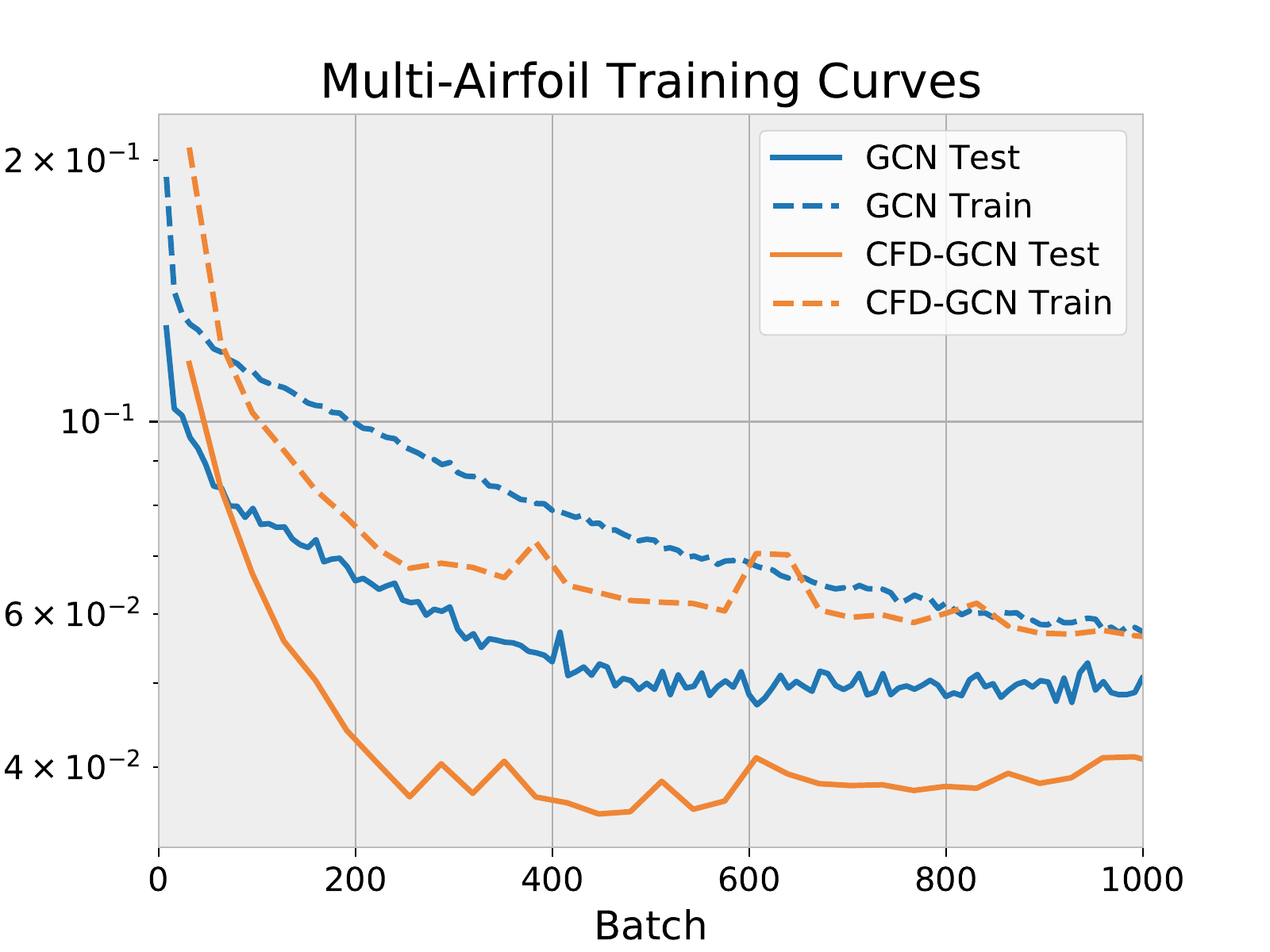}}
\else
    \centerline{\includegraphics[width=0.5\linewidth]{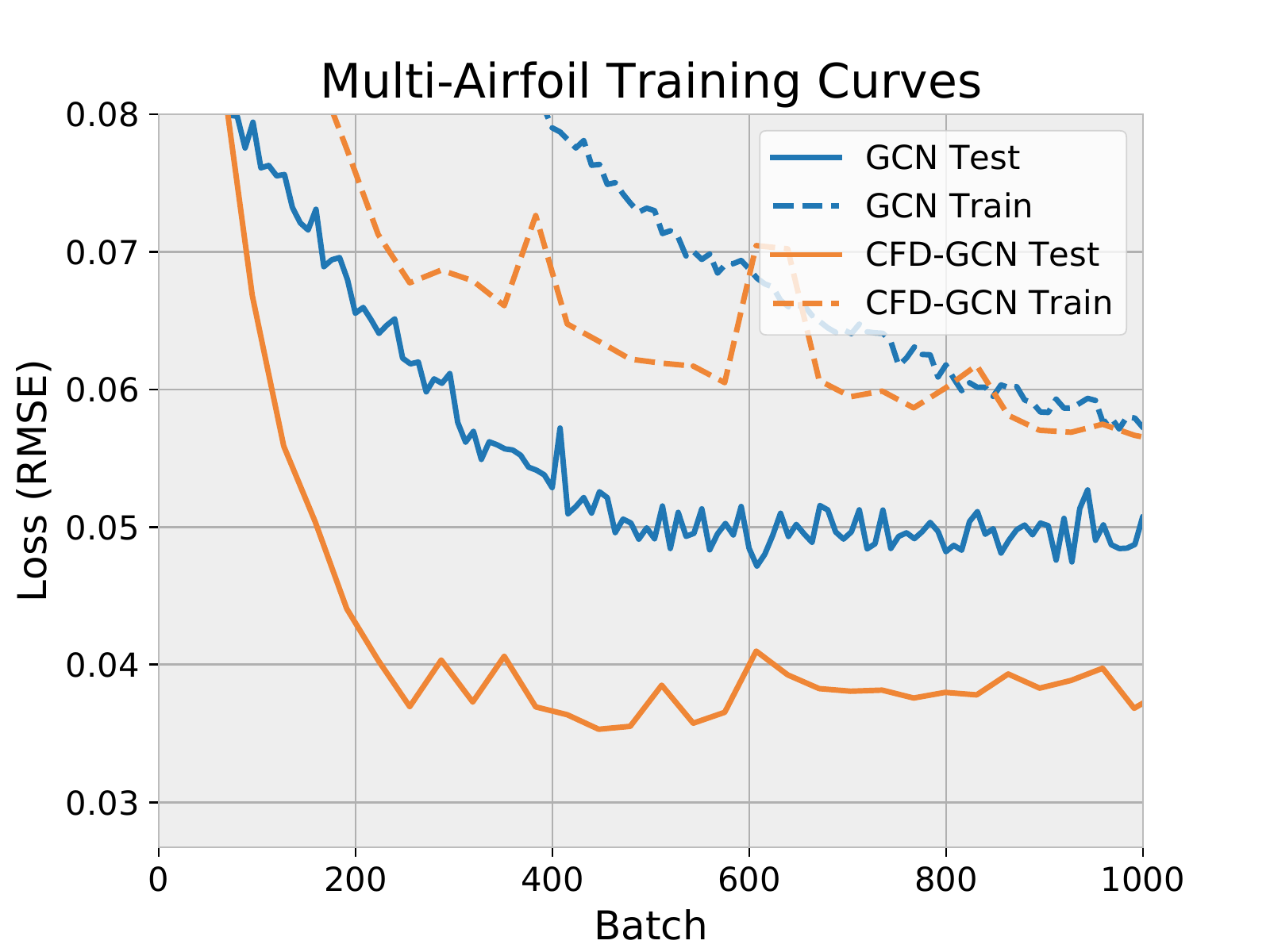}}
\fi
\caption{Training curves for the multi-airfoil experiment. As in the previous generalization task, the CFD-GCN is able to generalize better to the new conditions present in the test set. The test loss is lower than the training loss here simply because the training and test set are composed of different airfoils, with meshes containing different number of nodes, and thus their losses are not directly comparable.}
\label{fig:multi_curve}
\end{center}
\vskip -0.2in
\end{figure}

Figure~\ref{fig:multi_curve} presents the training curves for the CFD-GCN and the baselines.
As before, we observe that the CFD-GCN model generalizes better to the test set, which contains the previously unseen airfoil. With an RMSE of $0.34$ the upsampled coarse mesh baseline's results were too high to be displayed in the same plot. 

\newpage
\section{Mesh Optimization}

\begin{figure}[h!]
    \centering
    \begin{subfigure}[t]{0.475\linewidth}
        \centering
        \includegraphics[width=0.9\linewidth]{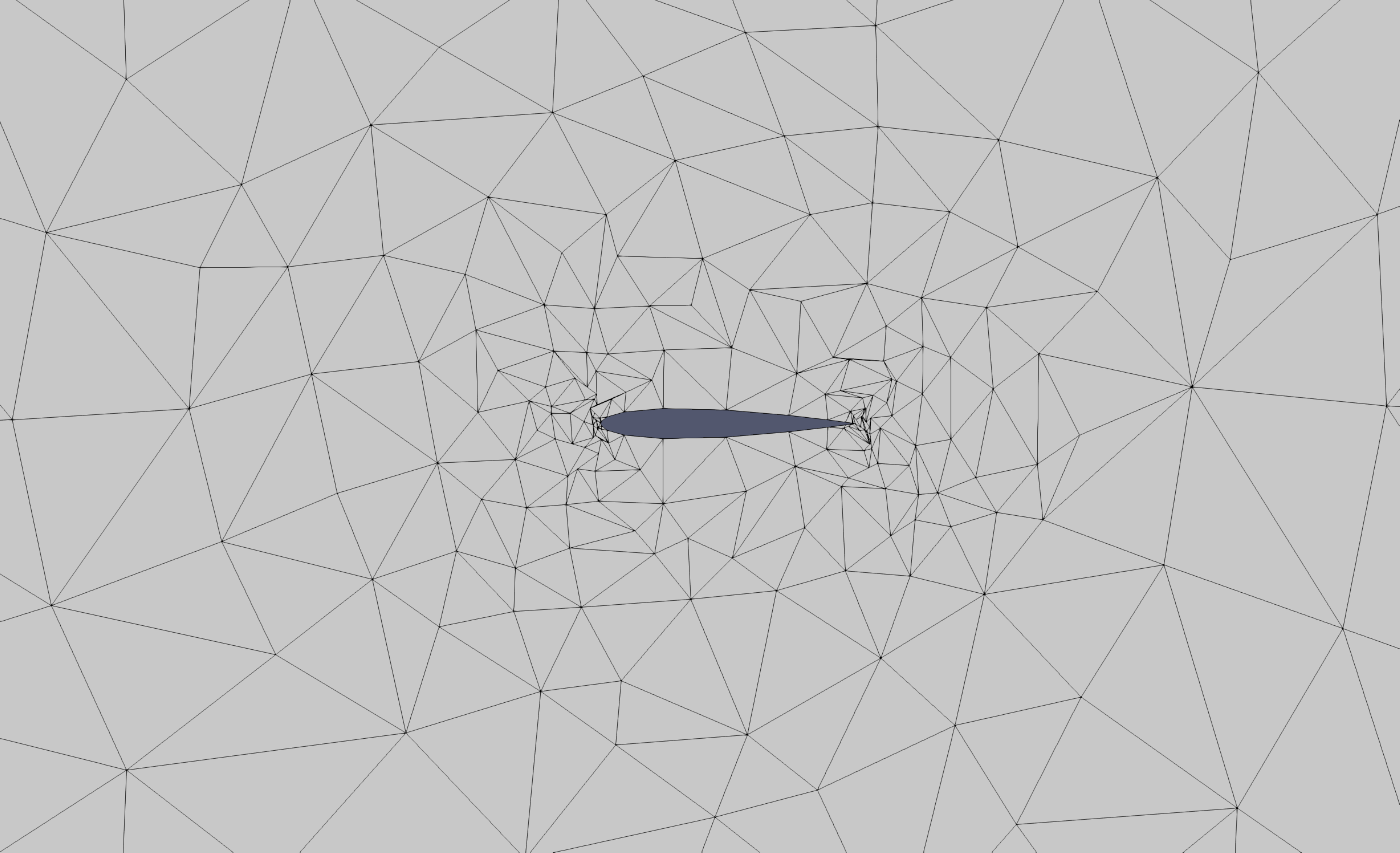}
        \caption{The NACA0012 mesh after training.}
    \end{subfigure}
    \begin{subfigure}[t]{0.475\linewidth}
        \centering
        \includegraphics[width=0.9\linewidth]{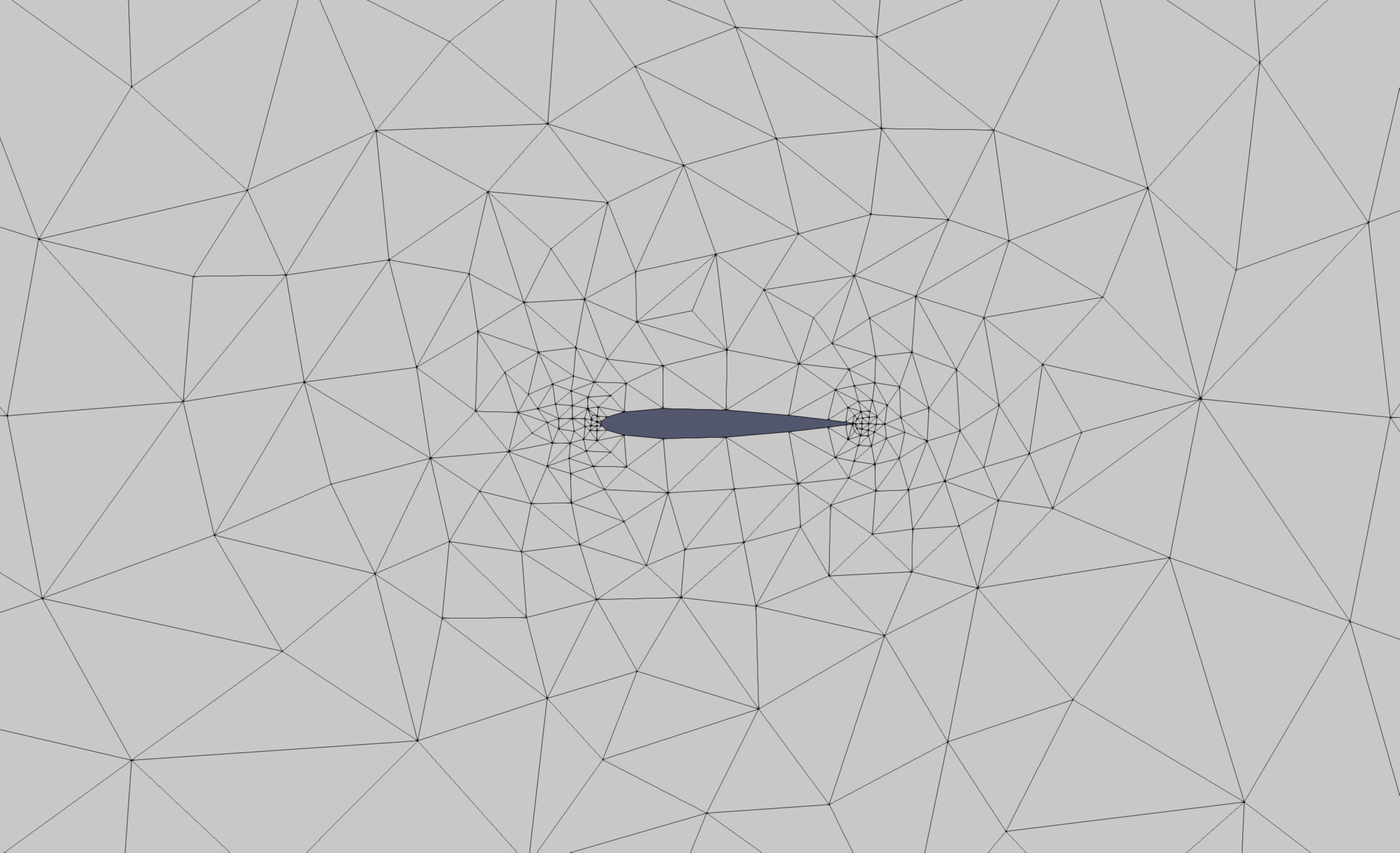}
        \caption{The original NACA0012 mesh before training.}
    \end{subfigure}
    
    \begin{subfigure}[t]{0.475\linewidth}
        \vskip 0.1in
        \centering
        \includegraphics[width=0.9\linewidth]{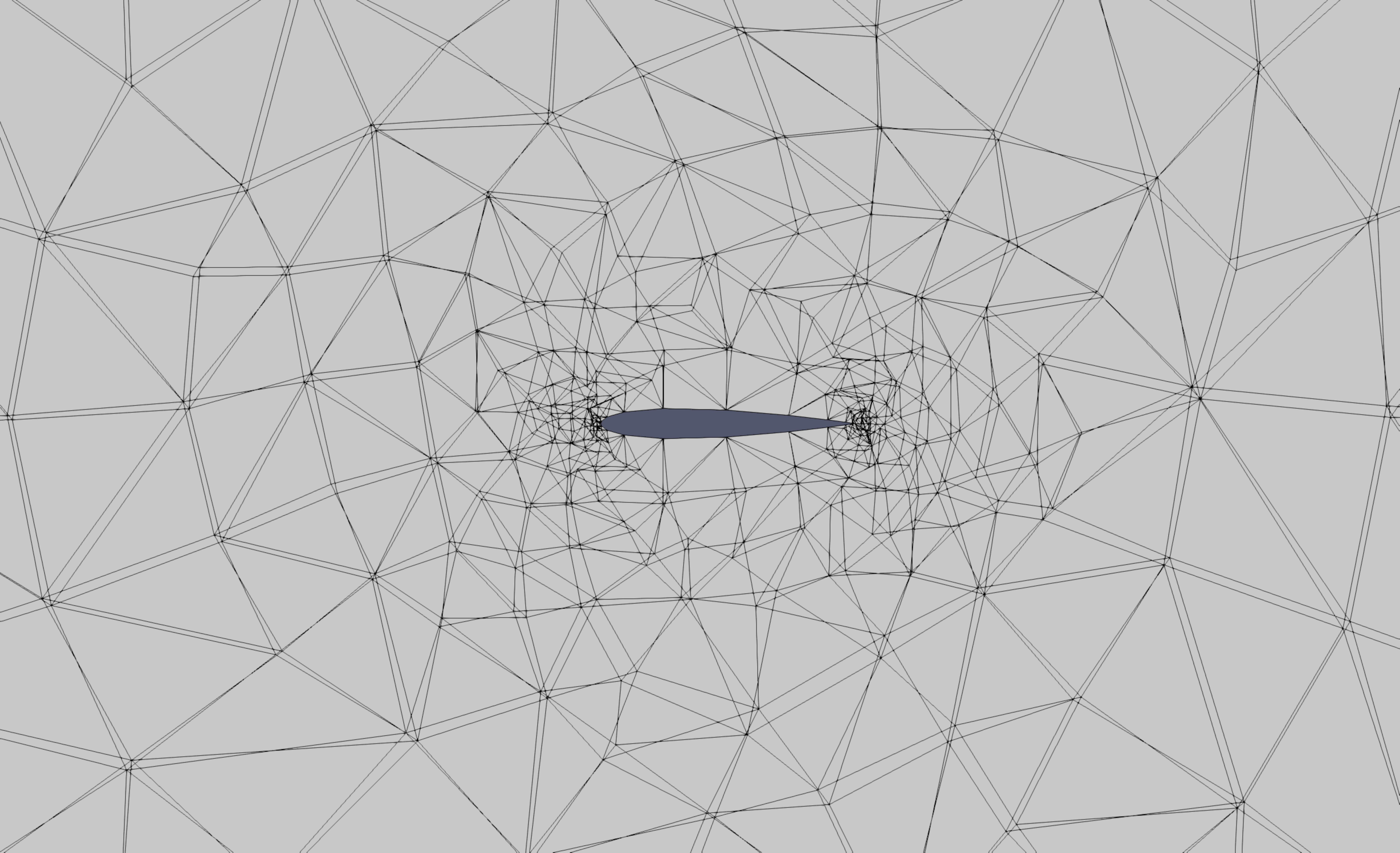}
        \caption{The meshes before and after training, superimposed.}
    \end{subfigure}
    \begin{subfigure}[t]{0.475\linewidth}
        \vskip 0.1in
        \centering
        \includegraphics[width=0.9\linewidth]{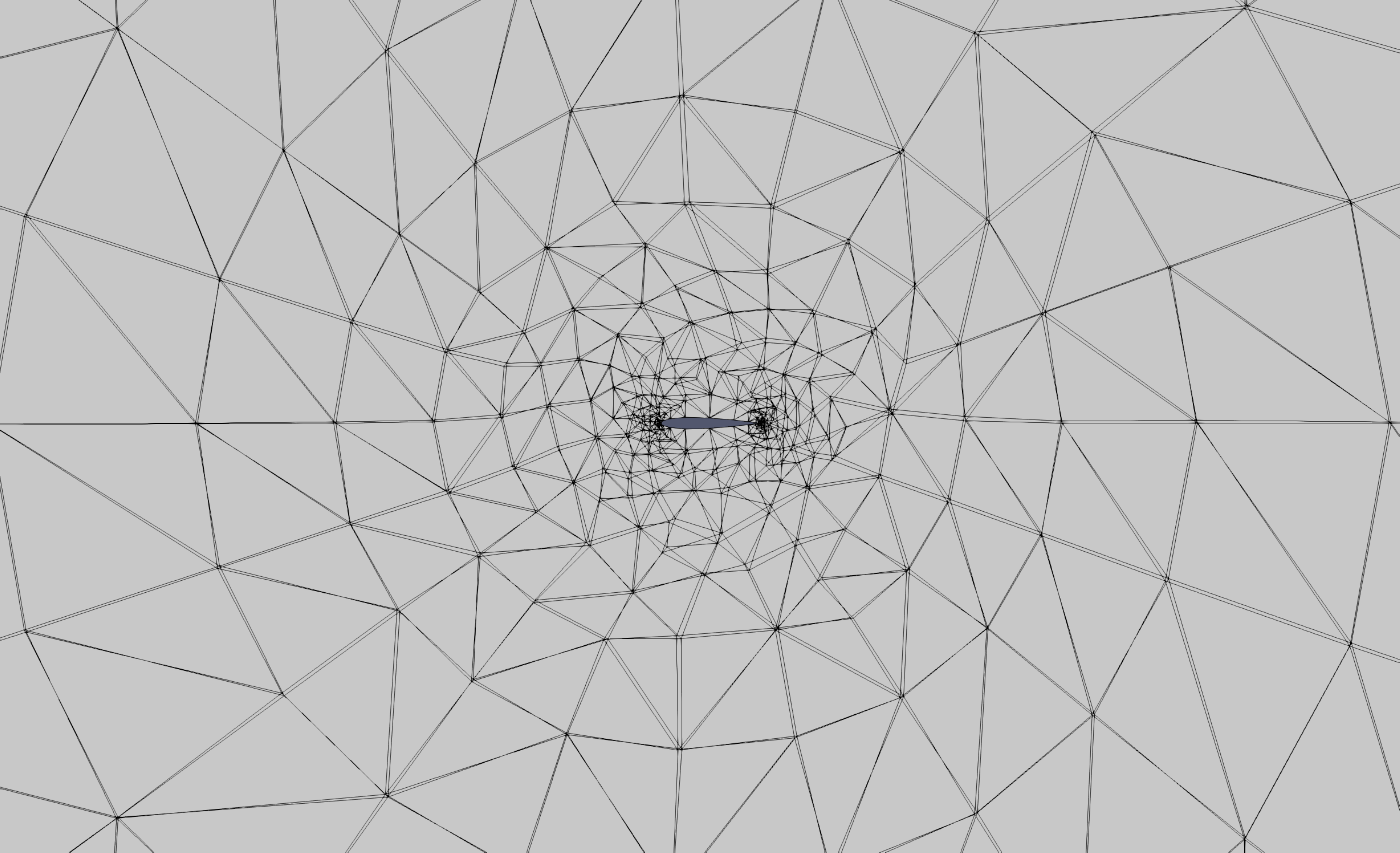}
        \caption{Farther view of before and after meshes, superimposed.}
    \end{subfigure}
    
    \caption{Comparison of the coarse mesh before and after being optimized during training. Changes are greater around the airfoil, where the gradients of the loss are large. Regions further away from the wing, which do not affect prediction strongly, are mostly unaltered.} 
    \label{fig:opt_mesh}
\end{figure}

In Figure~\ref{fig:opt_mesh}, we present changes to the mesh during the optimization process for the generalization task.

\newpage
\section{Interpolation Experiment}
\label{app:pred}

\begin{figure}[h!]
    \centering
    \begin{subfigure}{\columnwidth}
        \centering
        \includegraphics[width=0.75\columnwidth]{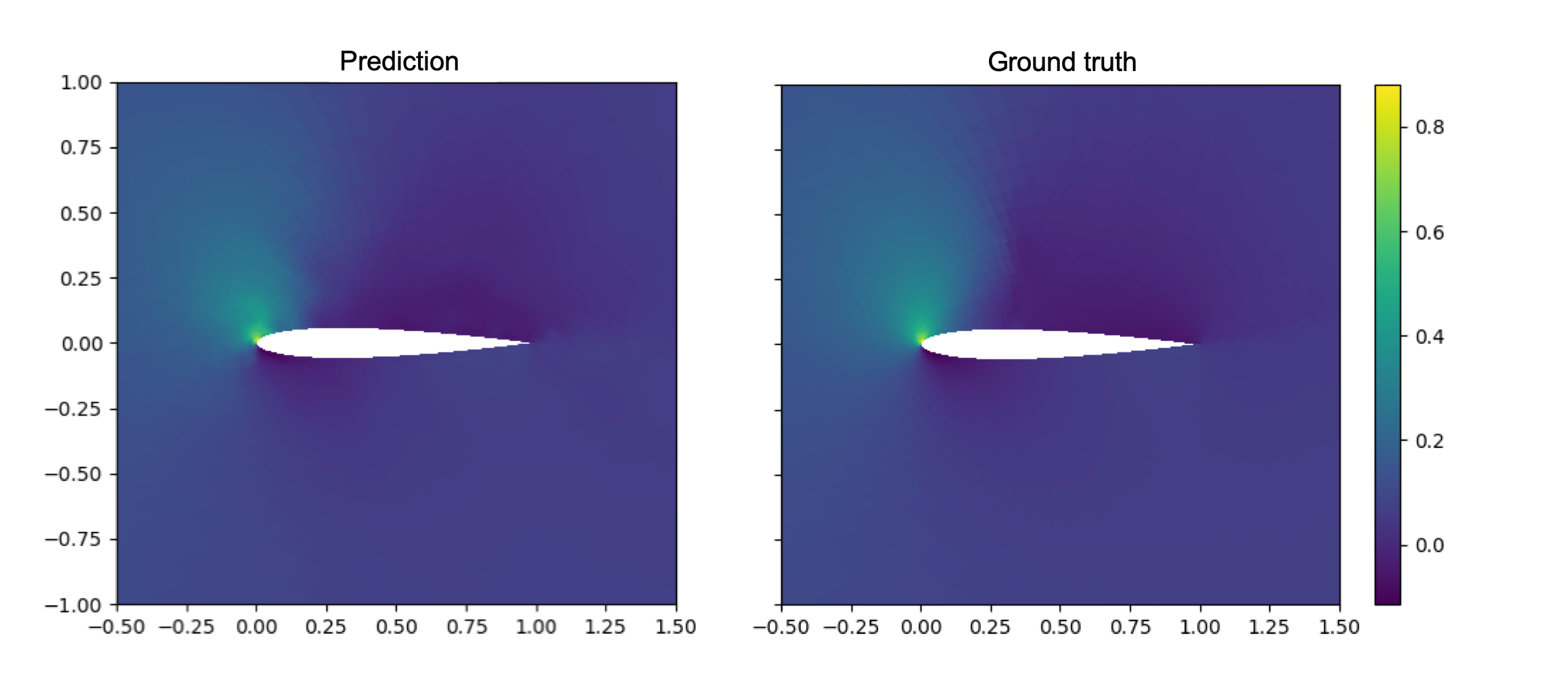}
    \end{subfigure}
    
    \begin{subfigure}[t]{\columnwidth}
        \centering
        \includegraphics[width=0.75\columnwidth]{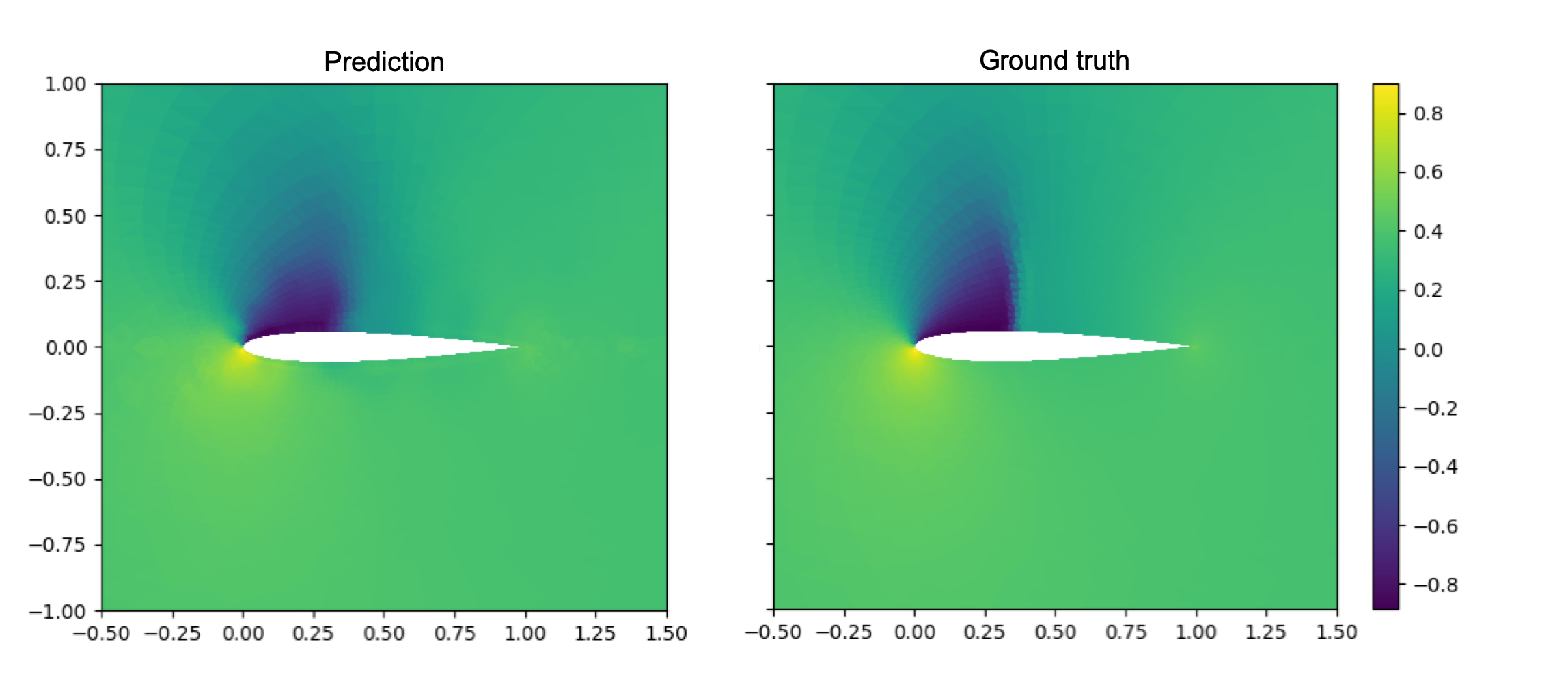}
    \end{subfigure}
    \caption{CFD-GCN model prediction and ground truth for a test sample in the interpolation task. The y component of the velocity and the pressure output fields for the same sample as in Figure~\ref{fig:pred} are presented here.}
    \label{fig:pred_rest}
\end{figure}

In Figure~\ref{fig:pred_rest}, we present the predictions and ground truth for the fields that were omitted in the main text for the interpolation task in Section~\ref{sec:pred}.

\newpage
\section{Generalization Experiment}
\label{app:gen}

\begin{figure}[h!]
    \centering
    \begin{subfigure}{\columnwidth}
        \centering
        \includegraphics[width=0.75\columnwidth]{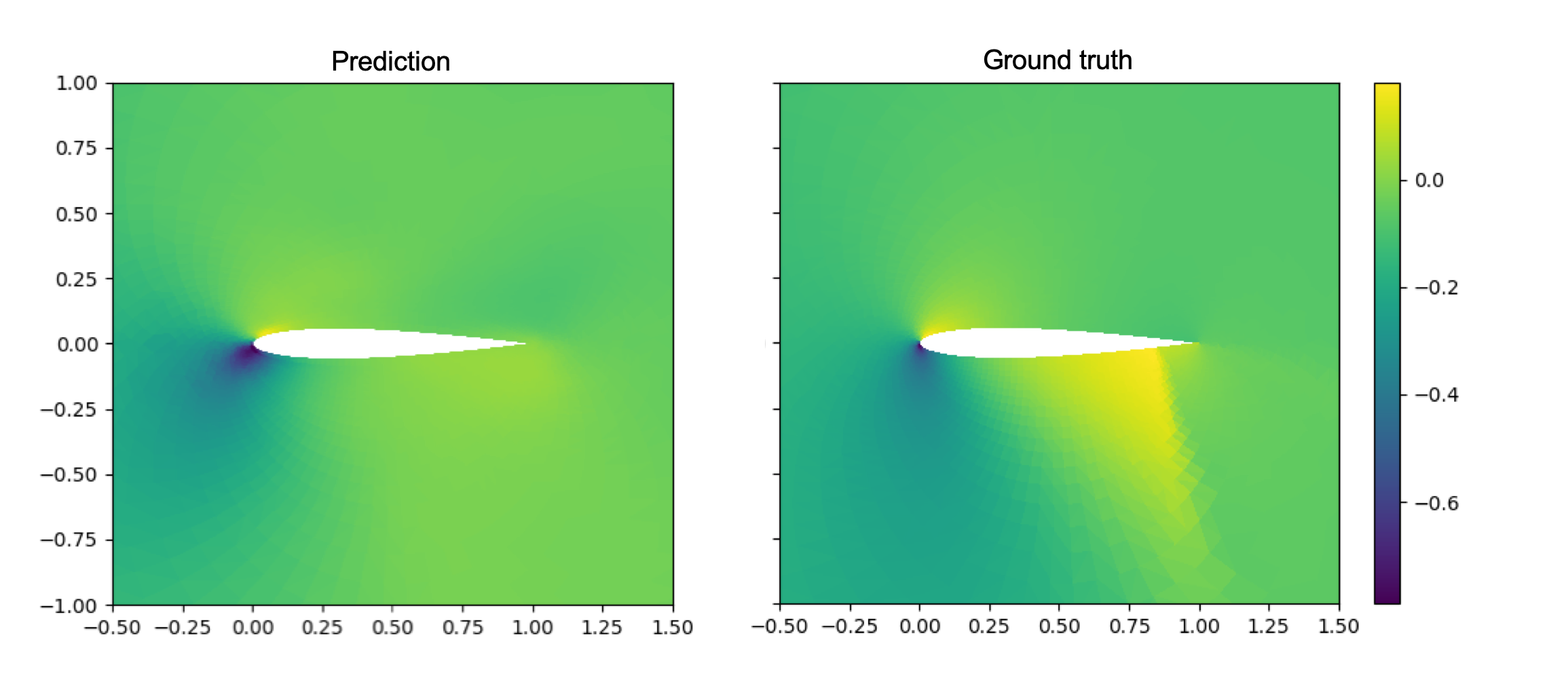}
    \end{subfigure}
    
    \begin{subfigure}[t]{\columnwidth}
        \centering
        \includegraphics[width=0.75\columnwidth]{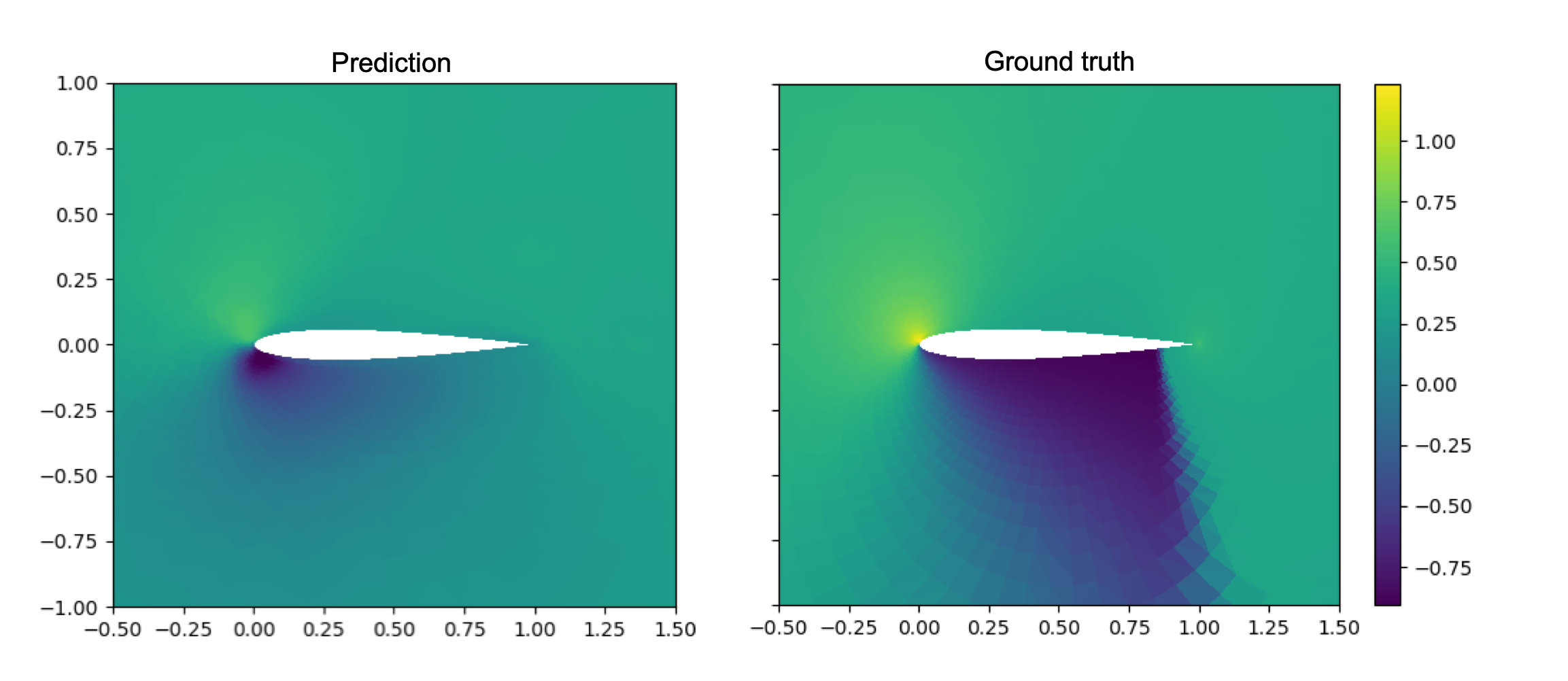}
    \end{subfigure}
    \caption{The GCN baseline prediction for a test sample with a large shock in the generalization task. The y component of the velocity and the pressure output fields for the same sample as in Figure~\ref{fig:gcn_gen} are presented here.}
    \label{fig:gcn_gen_rest}
\end{figure}

\begin{figure}[h!]
    \centering
    \begin{subfigure}{\columnwidth}
        \centering
        \includegraphics[width=0.75\columnwidth]{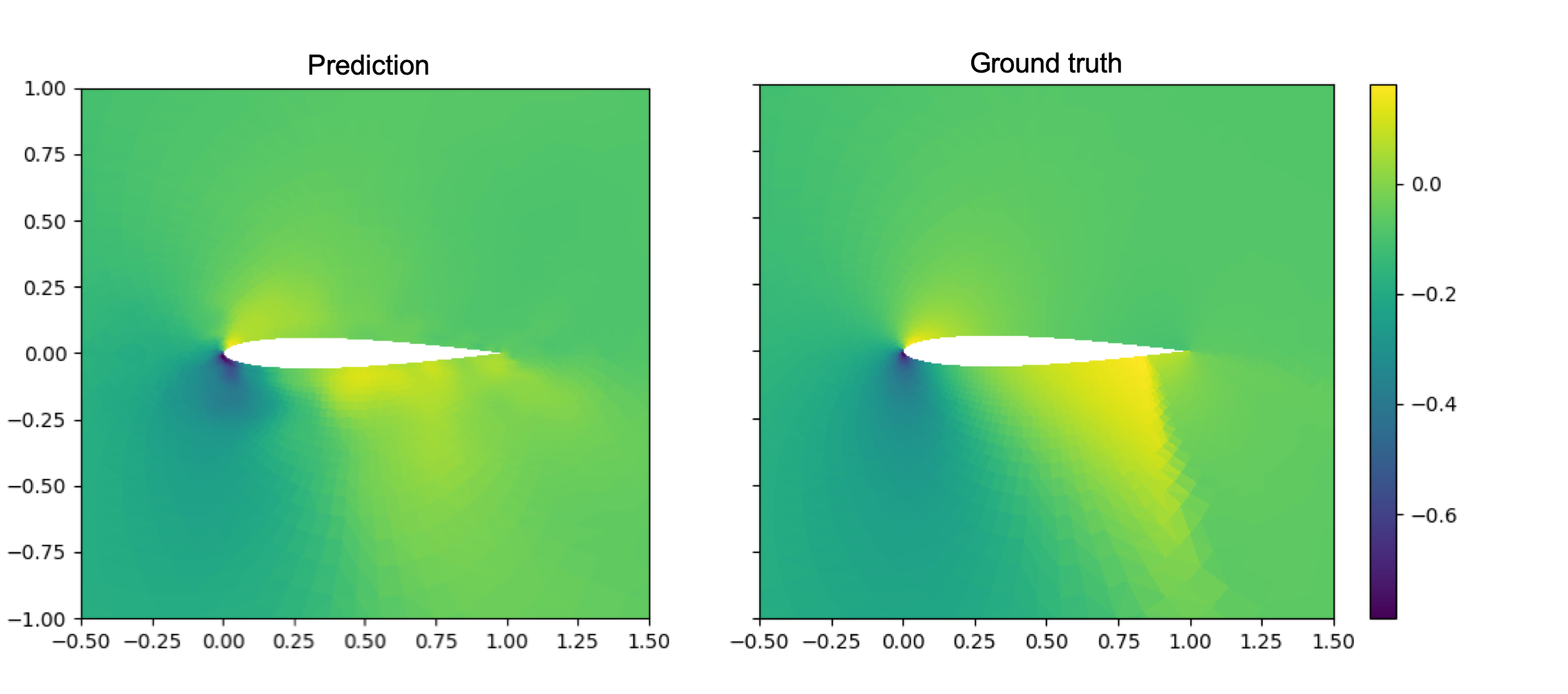}
    \end{subfigure}
    
    \begin{subfigure}{\columnwidth}
        \centering
        \hskip -0.067in
        \includegraphics[width=0.76\columnwidth]{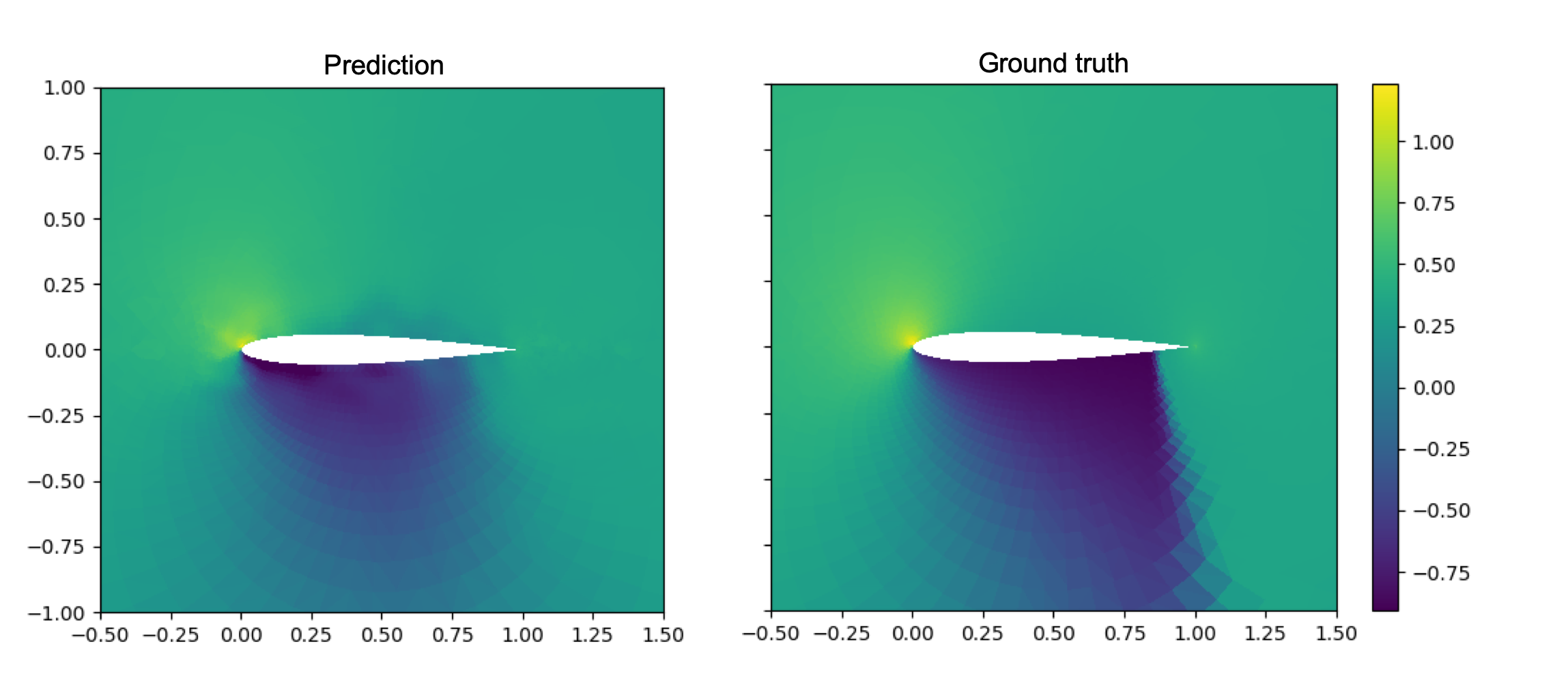}
    \end{subfigure}
    \caption{The CFD-GCN model prediction for a test sample with a large shock in the generalization task. The y component of the velocity and the pressure output fields for the same sample as in Figure~\ref{fig:cfd_gen} are presented here.}
    \label{fig:cfd_gen_rest}
\end{figure}

\begin{figure}[h!]
    \centering
    \begin{subfigure}{\columnwidth}
        \centering
        \includegraphics[width=0.75\columnwidth]{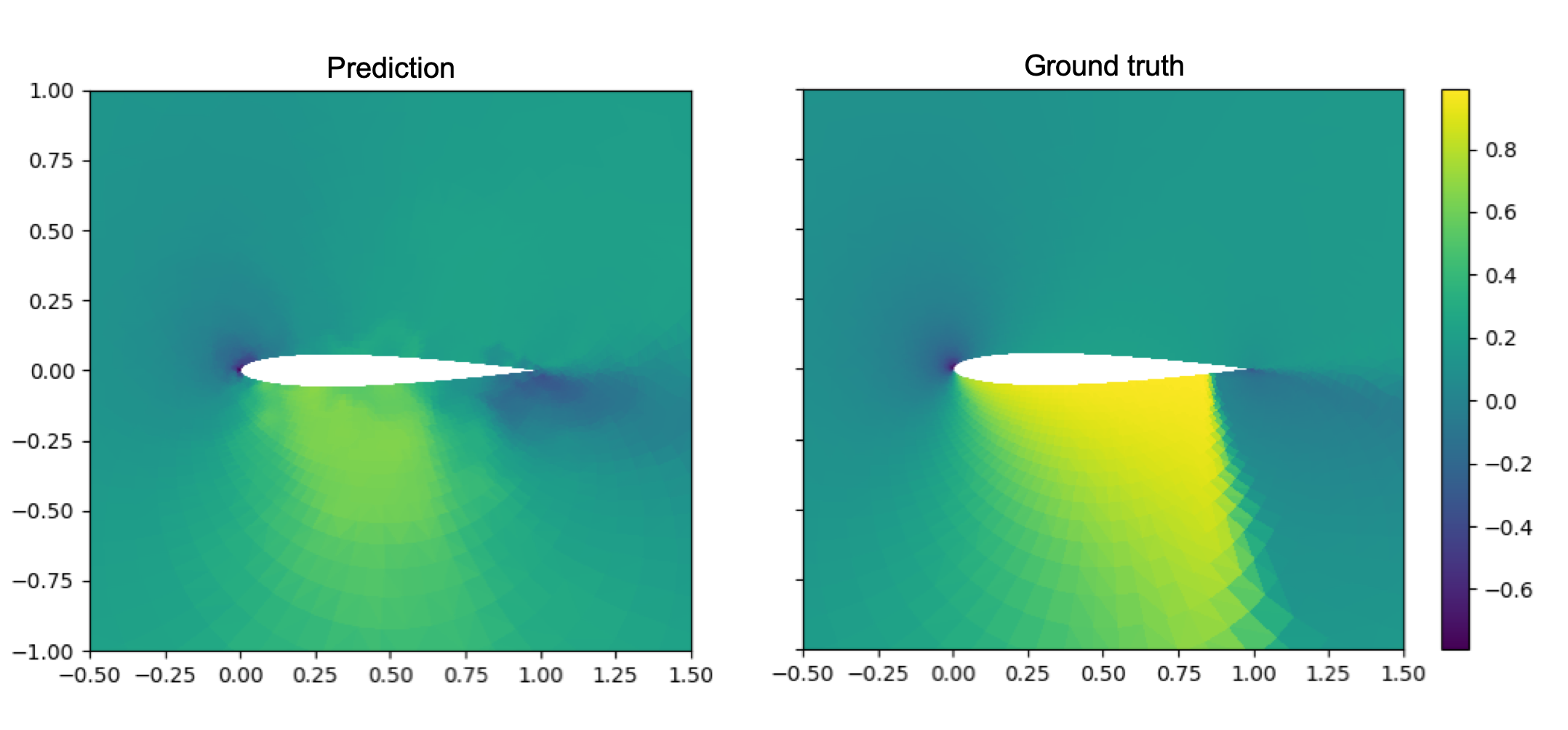}
    \end{subfigure}
    
    \begin{subfigure}{\columnwidth}
        \centering
        \hskip -0.067in
        \includegraphics[width=0.76\columnwidth]{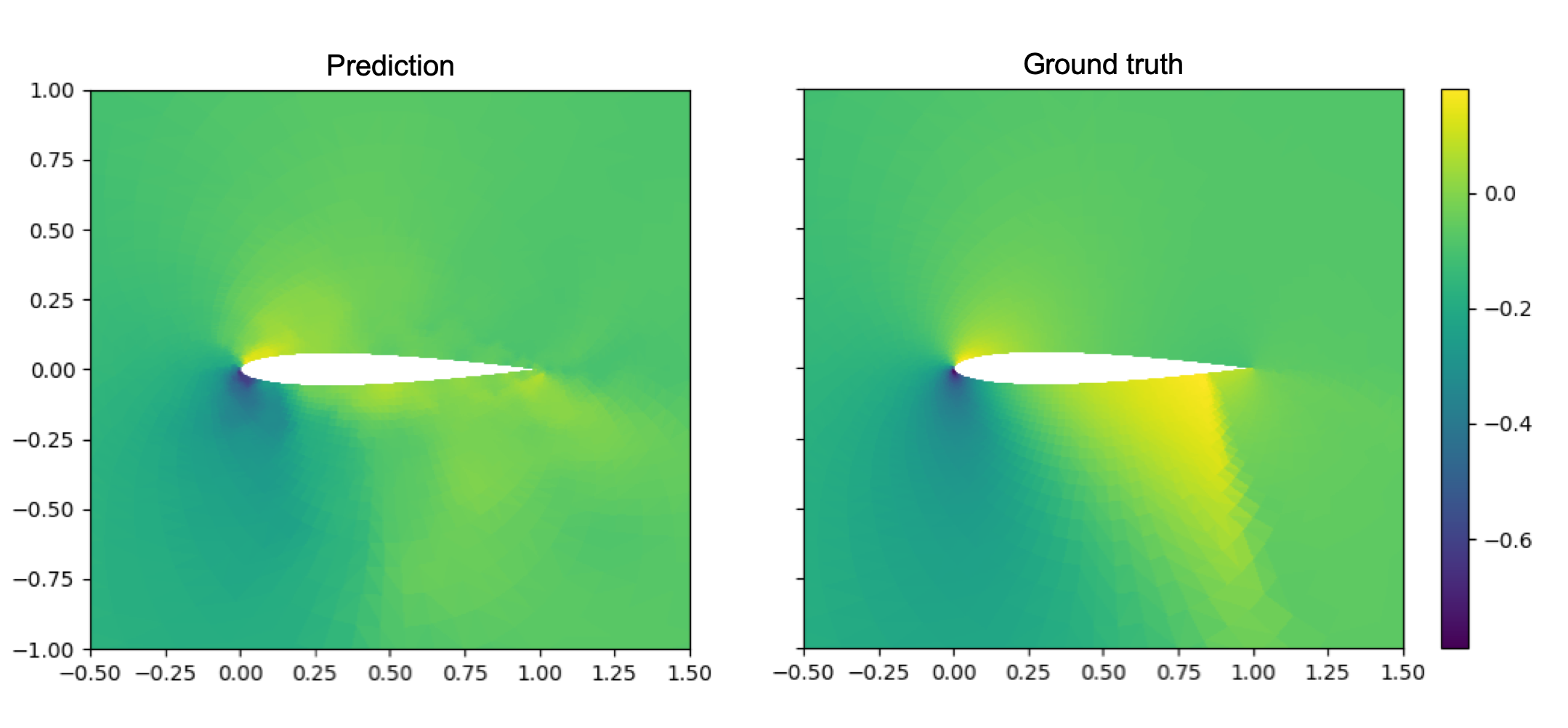}
    \end{subfigure}
    
    \begin{subfigure}{\columnwidth}
        \centering
        \hskip -0.067in
        \includegraphics[width=0.76\columnwidth]{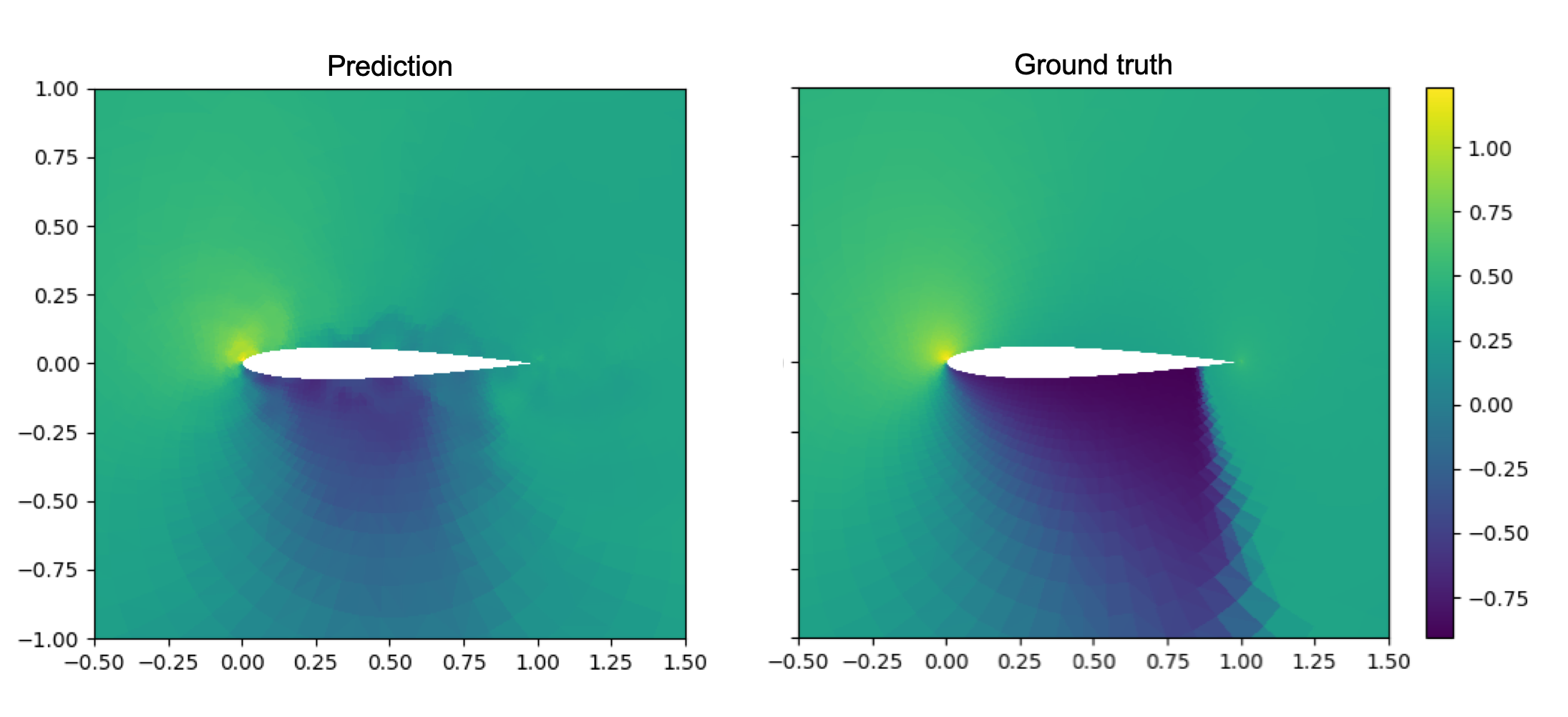}
    \end{subfigure}
    \caption{The upsampled coarse mesh baseline prediction for a test sample with a large shock in the generalization task. The x and y components of the velocity and the pressure output fields are presented here.}
    \label{fig:ucm_gen}
\end{figure}

In Figures \ref{fig:gcn_gen_rest} and \ref{fig:cfd_gen_rest}, we present the predictions and ground truth for the fields that were omitted in the main text for the interpolation task in Section~\ref{sec:gen}. In Figure~\ref{fig:ucm_gen}, we present a prediction for the upsampled coarse mesh baseline.

\clearpage
\clearpage

\end{document}